\documentclass{article}
\usepackage{iclr2027_conference,times}

\usepackage[utf8]{inputenc}
\usepackage[T1]{fontenc}
\usepackage{hyperref}
\usepackage{url}
\usepackage{booktabs}
\usepackage{array}
\usepackage{amsfonts}
\usepackage{amsmath}
\usepackage{amssymb}
\usepackage{nicefrac}
\usepackage[expansion=false]{microtype}
\usepackage{enumitem}
\usepackage{graphicx}
\usepackage{xcolor}
\hypersetup{colorlinks=true, linkcolor=blue!45!black, citecolor=blue!45!black, urlcolor=blue!45!black}

\newif\ifrevmarks
\revmarksfalse
\definecolor{revcolor}{HTML}{0050E0}
\DeclareRobustCommand{\rev}[1]{\ifrevmarks{\color{revcolor}#1}\else#1\fi}

\title{Steering Interference Reflects the Model's\\
Defaults, Not the Behavior Directions}

\author{Srikanth Malla\textsuperscript{*}, Chiho Choi \& Joon Hee Choi \\ Samsung Semiconductor US \\ \texttt{\{srikanth.m, chiho1.choi, jh4.choi\}@samsung.com}}

\iclrfinalcopy
\begin{document}
\maketitle
\lhead{}
{\renewcommand{\thefootnote}{}\footnotetext{\textsuperscript{*}Correspondence to \texttt{srikanth.m@samsung.com}.}}

\begin{abstract}
Activation steering promises modular control of language model behavior: a behavior
such as politeness corresponds to a direction in a model's activations, and adding
that direction while it generates should switch the behavior on and leave everything
else alone. It does not. We ask what decides which other behaviors move, and by how
much, and find that \rev{on average it points to the model} rather than the behavior being
steered. \rev{Steers mostly lower what the unsteered answer shows and raise a set of behaviors
the model falls back on (on Qwen2.5-7B, for example, chiefly what the judge scores as} refusal,
sycophancy, and poeticism\rev{)}.

Three results across 24 behaviors and ten instruction-tuned models support this, every
effect read off the generated text by a language-model judge rather than off a probe.
That readout matters: \rev{on Qwen2.5-7B} all 24 behaviors are linearly decodable, but only
20 change what the model writes. First, \rev{on average, directions} carrying no behavioral
content, \rev{of a real steer's size and layer, move} the same behaviors
in \rev{a similar} order as real steers do\rev{, clearly on five models, weakly on five}.
Second, \rev{much} interference runs one way, so it cannot be an overlap between two
directions: steering profanity makes the model toxic, while steering toxicity leaves
profanity untouched. Third, with a behavior held out entirely, geometry measured on the
others explains \rev{little} of the interference it takes part in. \rev{The one-way structure and
the failure of geometry hold on all ten models. Base models share the one-way structure, yet
only one of four has refusal or sycophancy among the five behaviors steers induce most, so
post-training appears to install them. Neither
geometry nor random directions predicts one pair of behaviors well. For single-layer
additive difference-of-means steering, this implies} that disentangling behavior directions
cannot by itself make steering modular.
\end{abstract}

\section{Introduction}
\label{sec:intro}

\begin{figure}[t]
\centering
\includegraphics[width=1.000\textwidth]{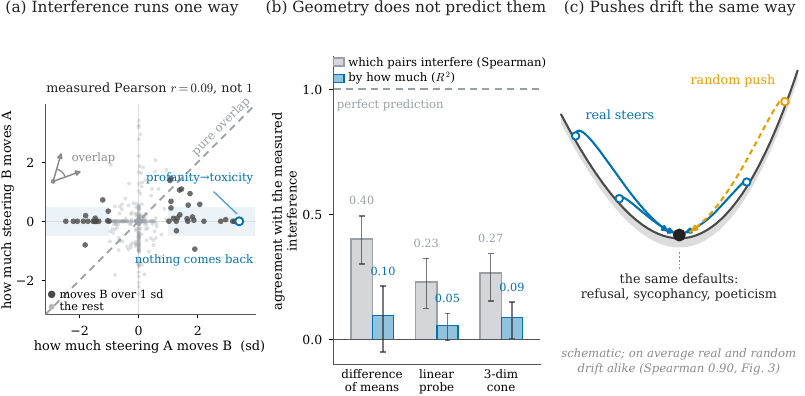}
\caption{\textbf{The paper's argument in one row: interference runs one way, geometry
cannot recover it, and \rev{on average pushes land} in the same place.} \textbf{(a)} Steering
one behavior moves others, and the effect runs one way. A and B are any two of the 20 steerable behaviors,
each point one ordered pair: how far steering A moves B against how far steering B
moves A, in units of the pooled standard deviation of the judge's score. \rev{Dark points are
pairs where steering A moves B by more than 1 sd, and the blue band is where the reverse
effect stays under 0.5 sd.} \rev{A geometric account}
treats interference as the overlap between two behavior directions (inset), one number
per pair, which forces the two equal and puts every point on the dashed line. Measured
the two correlate at Pearson $r=0.09$; 43 of the 380 effects exceed 1 sd and three in
four of those come back under 0.5 sd. \textbf{(b)} Nor is that structure recoverable
from the directions.
The two bars ask whether geometry ranks the interfering pairs correctly (Spearman) and
whether it gets their size right ($R^2$), 1.0 perfect on either; each takes the \rev{best}
predictor from Table~\ref{tab:predictor}.
\textbf{(c)} What we find instead, as a schematic: \rev{on average real steers and random
directions of the same size drift toward the same defaults, clearly on half the models.}}
\label{fig:concept}
\end{figure}

A transformer computes by carrying a running vector, the \emph{residual stream},
through its layers, each layer reading from it and writing back. Activation steering
exploits this directly: add a fixed direction to that stream while the model
generates, and the output shifts toward a target behavior. The technique needs no
fine-tuning and has become a standard tool for controlling refusal, sentiment,
sycophancy, and style at inference time
\citep{turner2023activation,rimsky2024caa,zou2023representation,arditi2024refusal}.
Because it is cheap and reversible, steering is increasingly
proposed as a lightweight control surface for alignment and safety. Its value rests on
one assumption, modularity: that behaviors can be toggled on demand without disturbing
the rest.
In practice this modularity leaks. Steering one behavior reliably moves others, so
inducing a desired trait quietly degrades or inflates unrelated ones
\citep{steeringsafety2025,pitfalls2026,unreliability2025}. A safety edit that
silently shifts unrelated behaviors is not a control one can trust, and what governs
that interference is poorly understood.

This paper asks what decides which other behaviors a steer moves, and finds that \rev{the
average pattern of side effects points to} the model rather than the behavior steered.
\rev{On average, random directions of a real steer's size and layer lower and raise the same
behaviors as real steers do, clearly on half the models. For a single pair of behaviors,
neither that nor geometry predicts much.}

The alternative, and the account recent work adopts, is geometric. If each behavior is
a direction, two behaviors should interfere to the extent that their directions point
the same way, which is what a cosine measures, and recent work formalizes this as
interacting subspaces coupled through shared decision pathways
\citep{lowrank2026,personality2026}. It is attractive because it is predictive in
principle, and falsifiable in two ways Sec.~\ref{sec:background} makes precise: overlap
is one number per pair, so the account predicts that two behaviors disturb each other
equally, and it is worth having only if it predicts for a behavior it was not fit on. We test both forms and both fail, which is what sends the explanation to the model
side.

Answering the question is harder than it looks, for two reasons that shape everything
below.

First, the answer cannot be read off the model's internals. Suppose we measured how much
steering behavior $i$ moved behavior $j$ by projecting the activations onto $j$'s
direction. Steering adds $i$'s direction to those activations, so that projection rises
in proportion to the overlap between the two, whether or not a word of the output
changed. Such a measurement confirms the geometric account whatever the model does, so
every effect has to be read from the generated text instead, which means a judge in the
loop and a steering coefficient calibrated per behavior.

Second, geometry cannot be tested on the behaviors it was fit on. The test is a
prediction: from the geometry of two behaviors' directions, predict how far steering one
moves the other, fitting on some pairs and checking on pairs the fit never saw. Which
pairs are held back decides what the check is worth. Each of the 20 steerable behaviors
appears in 38 pairs, 19 as the behavior steered and 19 as the behavior measured. So
hiding one pair, say the effect of steering profanity on toxicity, still leaves toxicity
in 37 others. A predictor can then get the
hidden pair right simply by having learned that toxicity is easily induced, with the
geometry contributing nothing. Only removing a behavior together with all 38 of its pairs asks whether geometry
says anything about a behavior it has never seen, and that is the test prior work does
not run.

We rebuild the measurement on those two changes, and run it over 24 behaviors spanning
register, affect, interpersonal stance, and content style, listed with their definitions
and prompt sets in Appendix~\ref{sec:app-behaviors}.

What the rebuilt measurement shows is the model side of the ledger, and one pair makes
the case. Steering profanity raises the model's toxicity by 3.4 standard deviations;
steering toxicity leaves profanity exactly where it was. Swearing reads as hostile, but
hostility does not make the model swear, and no quantity that is the same in both
directions can express the difference (Fig.~\ref{fig:concept}a).

\paragraph{What is new.}
\rev{That steering has side effects is known, and so is the fragility of post-trained behavior
under perturbation \citep{roguescalpel2025,ids2025,emergentmisalignment2025,assistantaxis2026}.
What has been missing is a measurement that separates a geometric account from a model-side
one. We read side effects from generated text and test geometry on behaviors it was not fit
on, where it explains at most about 15 percent of the variance on any model, against random
directions, which on average reproduce the pattern clearly on five of ten models and weakly
on the rest. Neither predicts
single pairs well, so a low cosine between two directions is no assurance that steering one
leaves the other alone.}

\paragraph{Contributions.}
\begin{itemize}[leftmargin=1.5em,itemsep=1pt,topsep=2pt,parsep=0pt]
\item \rev{\textbf{A measurement protocol.} Effects are read from generated text, with a judge in the loop and a coefficient calibrated per behavior, for 24 behaviors on ten models. On Qwen2.5-7B, the model we analyze in most detail, all 24 directions separate held-out responses, yet only 20 change what the model writes (Sec.~\ref{sec:map}).}
\item \rev{\textbf{A held-out test of geometry.} Holding out a behavior with all its pairs separates what geometry predicts from what a predictor learns from behavior names, and geometry explains little (Sec.~\ref{sec:geometry}).}
\item \rev{\textbf{A random-push contrast.} The average of random directions of a real steer's size and at its layer moves the same behaviors in a similar order as real steers (rank correlation 0.90 on Qwen2.5-7B), clearly on five of ten models and weakly on the rest (Secs.~\ref{sec:attractor} and~\ref{sec:scaling}).}
\item \rev{\textbf{The shape of the side effects.} Much interference runs one way, which no symmetric quantity can express, and one ordering of behaviors explains 0.39 of that part (Sec.~\ref{sec:hierarchy}).}
\end{itemize}

Sections~\ref{sec:map} to \ref{sec:scaling} run in the order the argument needs rather
than the order above: what the interference is, what shape it has, \rev{what a random push does to it,
whether geometry predicts it,} and how far it generalizes.

\section{Behavior directions, steering, and what geometry predicts}
\label{sec:background}

This section defines the three things we measure: a behavior direction, a steer, and
the matrix of cross-behavior effects. It then states the hypothesis we test.

\paragraph{A behavior direction and steering it.}
Take pairs of assistant responses that share a user turn and differ only in whether
the response expresses some behavior, say politeness. Record the residual stream for
each response. Subtract the mean over the responses without the behavior from the mean
over those with it. The difference points from the first group toward the second. That
vector is the behavior's \emph{direction}, one arrow in activation space per behavior. To steer the behavior, add its direction back into the
residual stream at one layer while the model generates, scaled by a multiplier we call
the \emph{steering coefficient}. The premise of the technique is that this switches the behavior on
and leaves the rest of the output where it was.

\paragraph{Interference and the matrix that records it.}
\rev{In practice,} steering behavior $i$ also changes how strongly the model expresses
behavior $j$. \rev{We record that in a square matrix $M$, the \emph{interference matrix}. A judge
rates each generated response on every behavior from 0 to 10 (Sec.~\ref{sec:method}). With
$\bar r_j^{(i)}$ the mean rating of $j$ over 32 neutral prompts while steering $i$,
$\bar r_j^{(0)}$ the same without steering, and $\sigma_j$ the standard deviation of $j$'s
ratings pooled over all conditions,
\begin{equation*}
M_{ij} = \bigl(\bar r_j^{(i)} - \bar r_j^{(0)}\bigr) / \sigma_j .
\end{equation*}}
Its diagonal is each behavior's effect on itself. Everything off the diagonal is a \emph{side
effect}, and the paper asks what determines those entries.

\paragraph{The geometric hypothesis and its two commitments.}
\rev{The hypothesis we test} is that $M_{ij}$ depends only on how much directions $i$ and
$j$ overlap, measured by the cosine of the angle between them. That commits to two
things, and we test both.

\emph{It must be even-handed.} The angle between two directions is one number, and it
does not change with which direction is named first. So $\cos(i,j) = \cos(j,i)$, and
the account predicts that $i$ disturbs $j$ exactly as much as $j$ disturbs $i$. To
measure the departure from that, split $M$ into the part that is equal both
ways, $S = (M + M^{\top})/2$, and the part that runs one way,
$A = (M - M^{\top})/2$; every square matrix splits this way uniquely.
Entries of $A$ are cases like our running example: steering profanity makes the model
toxic, but steering toxicity does not make it swear. Any interference recorded in $A$
is outside a geometric account in principle, not merely fitted badly by it. That makes $A$
the cleanest place to check the account, \rev{provided it is not noise (Sec.~\ref{sec:hierarchy})}.

\emph{It must transfer.} A geometric account is useful because the directions,
measured once, say which behaviors will interfere without the steer ever being run. It
has that use only if it also holds for behaviors the geometry was not fit on. So
we predict interference from overlap, fitting on some behaviors and testing on one held
out together with all 38 of its pairs.

\section{Method}
\label{sec:method}

Every number in the paper comes out of one pipeline: extract a direction for each
behavior, calibrate how hard to steer it, steer it and read the effect on every other
behavior from the generated text, and repeat over all ordered pairs to fill the matrix
$M$ of Sec.~\ref{sec:background}. Each step below names the section that reports it.
We present the pipeline on Qwen2.5-7B-Instruct (28 layers, hidden size 3584) and repeat
it unchanged on nine further models. \rev{We fixed every setting on it first, and at 7B it runs on
one GPU, so we also run the costlier analyses of Step 5 on it.} Metrics are defined with worked interpretations in
Appendix~\ref{sec:app-metrics}, the behaviors in Appendix~\ref{sec:app-behaviors}.

\paragraph{Step 1, a direction per behavior, and whether it steers (Sec.~\ref{sec:map}).}
We form each behavior's direction as in Sec.~\ref{sec:background}, from 12 contrastive
prompt pairs, once per layer, and score it on pairs it was not fit on: \rev{we fit it on four of
five folds, project the held-out responses onto it, and measure how well the projection
separates responses with and without the behavior} by the area under
the ROC curve (AUROC), where 1.0 is perfect and 0.5 is chance, against the same test with
the labels shuffled (Appendix~\ref{sec:app-directions}). We then calibrate one steering
coefficient per behavior against a language model judge, rather than fix one for all.
\rev{The judge is the model under study, queried without steering; it reads each response alone
and rates all 24 behaviors in one JSON reply. A fixed Qwen2.5-72B judge reproduces the matrix
structure on the three models tested (Appendix~\ref{sec:app-judge-ablation}).}
Over a grid of coefficients we generate on 32 neutral prompts, have the judge rate every
generation on all behaviors from 0 to 10, and measure what fraction of adjacent word
pairs are distinct, which collapses when the model repeats itself. A behavior's
coefficient is the one at which the judge's rating of that behavior reaches a target
level while that fraction stays above a floor \rev{of 0.6, which removes coefficients, never single
generations}; a behavior with no such coefficient is
dropped (Appendix~\ref{sec:app-judge}). \rev{The floor sees repetition only, so
Appendix~\ref{sec:app-checks} repeats the analysis without degenerate text.}

\paragraph{Step 2, the interference matrix (Sec.~\ref{sec:map}).}
Entry $M_{ij}$ is the change in the judge's rating of $j$ when steering $i$ at its
calibrated coefficient, against unsteered baselines on the same prompts. We divide each
column of $M$ by a standard deviation pooled across conditions, so 1.0 means the steer
moved that rating by one typical spread. An entry is significant when a prompt bootstrap
excludes zero. \rev{These are strong steers: on Qwen2.5-7B the median steer adds about twice the
norm of the mean activation at its layer (Appendix~\ref{sec:app-checks}).}

\paragraph{Step 3, the structure of $M$ (Sec.~\ref{sec:hierarchy}).}
We decompose $M$ into a sum of simple patterns, ordered from the one explaining the most
to the least, and report the share the first few carry. We also fit one number per
behavior for how strongly it drives others and one for how easily it is induced, and
rank behaviors by the difference, which we call \emph{source strength}.

\paragraph{Step 4, the random push (Sec.~\ref{sec:attractor}).}
We repeat the measurement with \rev{24} random directions in place of behavior directions.
\rev{Each takes the largest of 0.5, 0.75, 1, 1.5 or 2 times a reference size near the median real
steer's that keeps the distinct-pair fraction above 0.9, or the smallest if none does, and
enters at layer 7 on every model (a \emph{layer-7 push}), the layer of 17 of the 20 real steers on Qwen2.5-7B. A push
is therefore comparable to a real steer in size rather than exactly matched: on Qwen2.5-7B
the sizes average 1.5 times that reference. Across models (Sec.~\ref{sec:scaling}) we use
\emph{matched pushes} instead: each copies the layer and size of one real steer, with no
selection on size, and its text is saved (Appendix~\ref{sec:app-new}).}

\paragraph{Step 5, geometry as a predictor, and acting on it (Sec.~\ref{sec:geometry}).}
We regress each off-diagonal entry of $M$ on 13 features of the two directions:
symmetric overlap, directional features that distinguish the two orderings, and, as a
deliberately generous addition, each direction's norm and each behavior's own on-target
effect. We fit both ridge regression and gradient-boosted trees, evaluated
by leave-one-behavior-out cross-validation that removes a behavior together with all 38
of its pairs, scored by held-out Spearman rank correlation and $R^2$ against a
label-shuffled null (Appendix~\ref{sec:app-analysis}). We then repeat the measurement
with \rev{two modified steers}: one with the components along every
other behavior's direction removed \rev{(orthogonalized), the remedy the geometric account
implies}, and one rescaled so the activations keep their
unsteered length \rev{(anchored), the remedy a model-side account suggests}.

\paragraph{Step 6, replication (Sec.~\ref{sec:scaling}).}
We run \rev{the pipeline} on ten instruction-tuned models spanning four families and 1.5B
to 72B parameters (Table~\ref{tab:scaling}); \rev{the probes, cones, and modified steers of Step 5
run on Qwen2.5-7B only}. Each model is steered and judged by itself,
so cross-model differences reflect the model under study and not a shared judge. The
layers searched, the coefficient grid, and all thresholds are held fixed at the
Qwen2.5-7B settings.

\section{Linear decodability does not imply steerability}
\label{sec:map}

On Qwen2.5-7B all 24 behaviors are linearly decodable, yet only 20 change what
the model writes, so a readout that never reads the text would have called all 24
controllable. Before asking what decides the side effects, the map itself has to be
trustworthy. Every direction reaches held-out AUROC 1.000, with label-shuffled
controls near chance (0.26 to 0.62, mean 0.50). The four exceptions under
judge-in-the-loop calibration are confidence, verbosity, humor, and
anthropomorphism, the abstract, non-lexical stances: no coherent coefficient makes the judge
read them in the text. The other 20 move their own rating, the
diagonal of $M$, by a median of more than two of the judge's ten points.

Interference among those 20 is pervasive. Of the 380 off-diagonal entries, 149
are significant, where 95 percent intervals would place about 19 by chance at
this many tests. The average side effect is about a fifth of the average
on-target one, 0.37 against 2.07 in pooled-standard-deviation units.

\begin{figure}[t]
\centering
\includegraphics[width=0.680\textwidth]{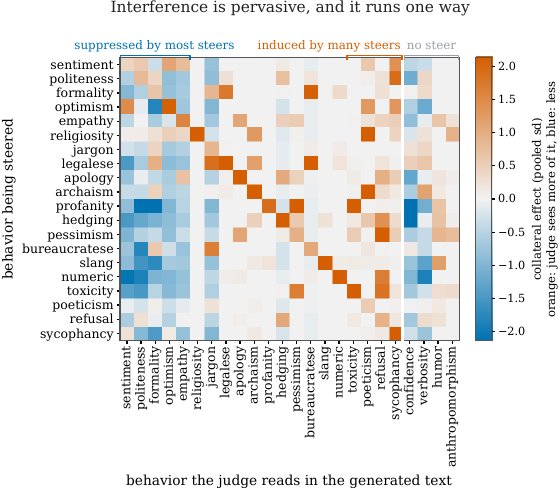}
\caption{\textbf{\rev{Across steers,} the same columns move: the prosocial columns on the
left fall and the sink columns \rev{next to the white line} rise, which a symmetric geometry cannot
express.} Judge-grounded interference
matrix over the 20 steerable behaviors. Each row is one steered behavior and each column
is one judge rating of the generated text, so the diagonal is its on-target effect and
everything off it is a side effect. \rev{Rows and the first 20 columns} use the same order, by source strength, a
behavior's drive on others minus how easily it is induced: resistant behaviors at the
top, sinks at the bottom. \rev{The last four columns have no row because no coherent coefficient
steers them, though other steers still move them.}}
\label{fig:heatmap}
\end{figure}

\section{\texorpdfstring{\rev{Much interference runs one way, and one ranking organizes part of it}}{Much interference runs one way, and one ranking organizes part of it}}
\label{sec:hierarchy}

Steering profanity moves toxicity by $+3.4$ pooled standard deviations; steering
toxicity moves profanity by $+0.0$. That asymmetry is the first of the geometric
account's two commitments to fail, and it is not one odd pair: across the matrix the
one-way part $A$ carries 0.68 of the total entry magnitude\rev{, or 0.46 on the squared scale
used below, close to the 0.5 that two unrelated effects would give}. It is visible pair by pair in
Fig.~\ref{fig:concept}a, where an effect above one pooled standard deviation
usually comes back under half of one, and as whole columns in
Fig.~\ref{fig:heatmap}. Overlap is even-handed by construction, so a geometric
account cannot express any of this: \rev{nearly half of the interference on that scale} is outside
its reach before any predictor is fit. \rev{Nor is $A$ noise: $A$ from one half of the prompts
correlates with $A$ from the other at 0.94, and at 0.91 to 0.96 across the ten models
(Appendix~\ref{sec:app-checks}).}

What does organize \rev{part of} it is a single ranking, from behaviors \rev{other steers lower} to
behaviors that are easily induced.

There is not much structure to organize: three patterns account for 0.71 of the
matrix's \emph{energy}, the sum of its squared entries. \rev{They carry about 0.40, 0.20 and 0.12. In the
first, almost every steer lowers politeness, sentiment and formality and raises pessimism,
refusal and toxicity; the second is a register pattern led by the legalese and formality
steers; the third moves mainly poeticism and sycophancy.} The 380 off-diagonal numbers
therefore come largely from three regularities rather than from
20 behaviors interfering each in its own way; the matrix is low-rank (confidence
intervals in Appendix~\ref{sec:app-analysis}). Much of that structure runs one
way: 0.46 of the energy sits in the one-way part $A$, which is the 0.68 above on
the squared scale, and a single pattern carries 0.54 of what is inside $A$. That
pattern is \rev{largely} the ranking\rev{, which on its own explains 0.39 of $A$}. Giving each behavior two numbers, how strongly it drives
others and how easily it is induced, already captures a third of the off-diagonal
magnitude; Table~\ref{tab:structure} in Appendix~\ref{sec:app-analysis}
decomposes it. The second number does most of the work: the matrix is mostly
about which behaviors are easy to induce, not which push hardest.

\begin{figure}[t]
\centering
\includegraphics[width=0.990\textwidth]{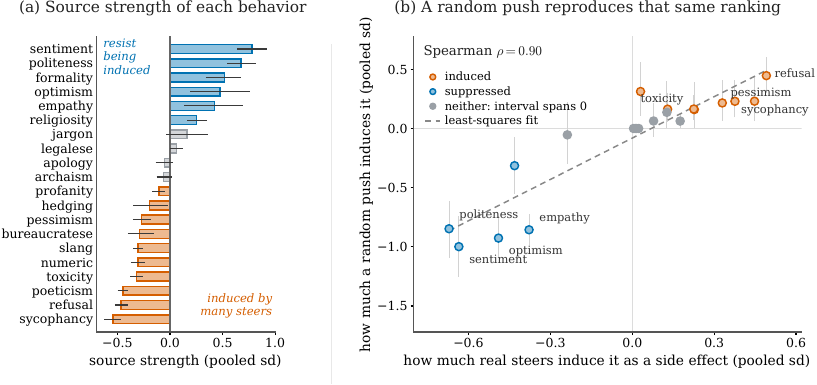}
\caption{\textbf{\rev{The interference ranking, and what a random push does to it, on Qwen2.5-7B.}}
\textbf{(a)} Every behavior ranked by source strength, its drive on others minus
how easily it is induced, with 95 percent bootstrap intervals: prosocial and
positive behaviors resist being induced (blue, top); failure, negative, and
filler behaviors are sinks (orange, bottom)\rev{; grey intervals span zero}.
\textbf{(b)} A random push, a direction \rev{of comparable size}, induces each behavior
(vertical) in proportion to how easily real behavior steers induce it \rev{as a side effect}
(horizontal\rev{; a behavior's own steer is excluded}), at rank correlation 0.90. \rev{It leaves
profanity, religiosity and slang at exactly zero}, so it is not simply moving everything.}
\label{fig:hier_attr}
\end{figure}

Which behaviors sit where is not arbitrary. With bootstrap confidence intervals,
ten of the twenty behaviors are significant sinks, \rev{with source strength below zero,} led by
sycophancy, refusal, and poeticism, \rev{the three that more than half of the other steers raise}.
Six significantly \rev{\emph{resist}, with source strength above zero: other steers mostly lower
them or leave them alone. They are} led by sentiment, politeness, and formality. The prosocial and positive
behaviors sit at the resistant end, the failure, negative, and filler behaviors at
the sink end (Fig.~\ref{fig:hier_attr}a).

\section{A random push reproduces the same ranking}
\label{sec:attractor}

\rev{On Qwen2.5-7B, the average of 24 directions} chosen with no reference to any behavior
reproduces that ranking at rank correlation 0.90. We steer along random directions \rev{of
comparable size (Sec.~\ref{sec:method}). All 24 passed the coherence bar and none collapsed
(Sec.~\ref{sec:scaling}), and a random eight of them already reach 0.87. Matched pushes agree
at 0.87 as well.} The behaviors
such a push induces track how easily real steers induce each one, at Pearson 0.92 on the
values as well as 0.90 on the ordering (Fig.~\ref{fig:hier_attr}b). \rev{The ordering is more than
headroom: among the 12 behaviors with an unsteered rating of exactly zero it is still 0.77.}
So the ranking is not a property of the
behaviors' representations, as it would be if the behaviors many steers induce simply
lay where many other behaviors' directions point. A random push significantly induces
ten behaviors, led by the sinks refusal, numeric, poeticism, and sycophancy, and
suppresses seven, five of the six resistant behaviors together with confidence and
verbosity. \rev{That agreement is with the average over steers: one real steer's side effects
have a median rank correlation of 0.57 with the average over random pushes and 0.46 with the
other real steers.}

The push is not simply inducing everything. It induces toxicity (0.16 in
pooled-standard-deviation units, 95 percent interval $[0.07, 0.28]$) but produces
profanity at exactly zero, as it does for \rev{religiosity and slang}.

The set a steer falls back on is not what the model does anyway. \rev{We call the
behaviors a random push significantly induces the model's \emph{default set}: what it falls
back on when disturbed.} On neutral prompts the fourteen behaviors that do not resist average 0.27
judge points out of 10, against 3.78 for the six that do. So \rev{the more a behavior
shows unsteered, the harder it is to induce} (rank
correlation 0.75, $n=20$, loosely, since twelve sit at zero). \rev{Refusal,} a behavior the model
never volunteers here, is what a random push induces most, \rev{at least as the judge scores it;
much of what it scores as refusal in side effects has no refusal wording
(Appendix~\ref{sec:app-checks}).}

\section{Geometry does not predict interference out-of-sample}
\label{sec:geometry}

The test is a prediction: geometry fit on the other behaviors has to predict a
held-out behavior's interference. If a steer's side effects are set by the model's
defaults, the steered direction should carry \rev{little} information about them, and it
\rev{carries little} (every variant is in Table~\ref{tab:predictor}; 380 ordered pairs). We score two ways, since they can disagree: rank correlation (Spearman)
asks whether the \emph{ordering} is right, the coefficient of determination $R^2$
whether the \emph{amount} is, with 1 perfect and 0 no better than guessing the average.

Geometry orders the interfering pairs only weakly. The best held-out rank correlation is
\rev{0.40}, above the label-shuffled null but far from determinative. It says almost nothing
about how large those effects are: no variant reaches a held-out $R^2$ of 0.10. Cosine
overlap on its own does as well as a predictor handed the full directional feature set.
Geometry barely explains the one-way part $A$, which is the property a predictive
account most needs. A directional feature built to capture $A$ reaches rank correlation
\rev{$-0.04$} over the 190 behavior pairs, counting each pair once, and the best geometric
predictor of $A$ that we found reaches only 0.16.

\rev{Holding out a behavior removes only part of the shortcut of Sec.~\ref{sec:intro}, since the other
behavior of each test pair stays in training: behavior names alone reach held-out Spearman
0.43. With both behaviors unseen the names carry nothing and cosine overlap keeps 0.21, so
geometry is a small real effect and most of the richer predictors' score was behavior identity
(Table~\ref{tab:holdout}, Appendix~\ref{sec:app-checks}).}

A richer estimate of the direction does not rescue it. A natural objection is
that one difference-of-means direction is a weak estimate of a behavior's subspace. That
is not the problem. We repeat the test with linear probes,
which unlike a difference of means account for how the activation dimensions vary
together, and with a three-dimensional \emph{concept cone} per behavior: \rev{the three leading
directions along which its paired differences vary around their mean}, so that a behavior is a
small subspace rather than one arrow. \rev{A probe exists only at a behavior's best layer, so for
probes we test cosine overlap alone. The cone contains the difference of means, and we also
test the two together.} The conclusion is unchanged (Table~\ref{tab:predictor}): \rev{neither
buys anything over cosine overlap between difference-of-means directions.}

\begin{table}[!htbp]
\centering
\caption{\textbf{No variant explains a tenth of the interference.} Leave-one-behavior-out
prediction of interference from direction geometry (20 behaviors, 380 ordered pairs\rev{, on a
second run of the pipeline that agrees with the first at $r=0.99$}).
Both scores are held out, and 1.0 is perfect on either. \rev{Ridge is linear regression with an
L2 penalty. The probe row predicts the matrix from steering along the probes.} ``All 13'' is the full feature
set of App.~\ref{sec:app-analysis}, which includes two \rev{kinds of feature} that are not geometry at
all: each direction's norm and each behavior's own on-target effect. Richer estimates and
a higher-capacity model do not rescue either score. The boosted model reaches in-sample
$R^2$ 0.66 against held-out \rev{0.096}, the signature of memorizing its training behaviors.}
\label{tab:predictor}
\begin{tabular}{llcc}
\toprule
direction estimate & predictor & which pairs (Spearman) & by how much ($R^2$) \\
\midrule
difference of means & cosine overlap, ridge      & \rev{0.24} & 0.090 \\
                    & all 13, boosted trees     & \textbf{\rev{0.40}} & \textbf{\rev{0.096}} \\
linear probe        & cosine overlap, ridge      & \rev{0.23} & \rev{0.055} \\
3-dim concept cone  & cone overlap, ridge       & \rev{0.18} & \rev{0.009} \\
                    & cone $+$ cosine, ridge    & \rev{0.24} & \rev{0.072} \\
\bottomrule
\end{tabular}
\end{table}

\rev{Neither remedy helps.} Neither modified steer reduces side effects, at each behavior's coefficient held fixed.
The side effects, in the matrix's pooled-standard-deviation units, are unmoved
(orthogonalized $-0.03$, $[-0.08, 0.02]$; anchored $+0.02$, $[-0.02, 0.06]$); both
merely weaken the steer, cutting on-target effect from 2.2 judge points to 1.2 and 1.8.
\rev{Anchoring was our own prediction, and it failed: side effect per unit of on-target effect
rises from 0.25 to 0.34, and to 0.37 when orthogonalized.}

\begin{figure}[t]
\centering
\includegraphics[width=0.886\textwidth]{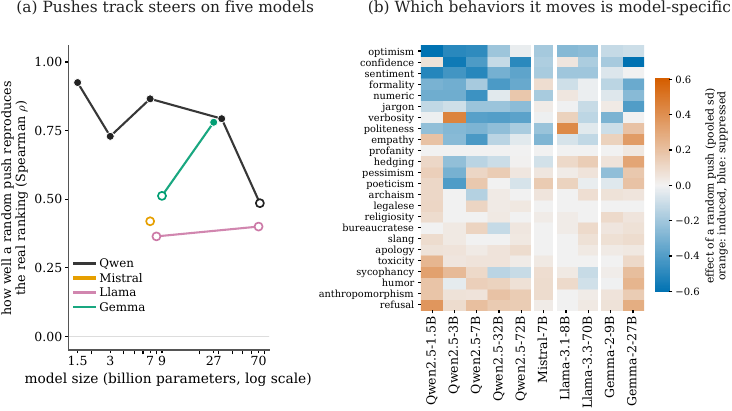}
\caption{\textbf{\rev{Random pushes track real steers clearly on five of ten models}; which
behaviors \rev{move differs by model}.} (a) \rev{Usable matched pushes}, directions
with no behavioral content \rev{that copy a real steer's layer and size, induce} the behaviors that real steers induce, in \rev{a similar}
order (lines connect a family by size\rev{; hollow points mark the five models whose 95 percent
interval reaches zero or nearly so, Sec.~\ref{sec:scaling}}). (b) What \rev{usable matched pushes do} to every behavior in every model, in pooled standard
deviations, drawn as in Fig.~\ref{fig:heatmap} and clipped beyond \rev{$0.6$}. Rows run from
behaviors a push suppresses \rev{across models} to those it induces; columns group by family
(white rules), by size within each. \rev{Suppression at the top holds across models; what
pushes induce is weaker and varies.}}
\label{fig:scaling}
\end{figure}

\section{The account across models and scale}
\label{sec:scaling}

Both central claims, \rev{the one-way structure} and geometry's failure to predict it,
hold on all ten models. Every model's interference matrix runs one way (one-way part 0.63 to
0.69 of total entry magnitude) and is low-rank (top three patterns 0.65 to 0.93 of the
energy, the squared scale). On all ten, geometry fails to predict it for a held-out
behavior (held-out $R^2$ $-0.11$ to 0.13\rev{, at most 0.15 for the best predictor on any model}). The pipeline is run unchanged from 1.5B to
72B across four families, each model steered and judged by itself
(Fig.~\ref{fig:scaling}; per-model values in Table~\ref{tab:scaling},
App.~\ref{sec:app-analysis}). \rev{Rebuilt without the degenerate text our checks detect, the one-way
and low-rank results barely move (Appendix~\ref{sec:app-checks}).} What shifts with scale is steerability: from 13 of 24
behaviors in the smaller models to all 24 at 70B, so the gap between what a probe
decodes and what a steer changes narrows.

\rev{Agreement with a random push is where the models differ. Some layer-7 pushes are invalid: they
fell below the coherence bar, or \emph{collapsed}, giving one identical set of ratings on nearly
all prompts, which the bar cannot detect; and on the largest models they enter far shallower
than real steers (Appendix~\ref{sec:app-checks}). Matched pushes remove both problems. Dropping
the few that are unusable, they agree at 0.73 to 0.93 on four of the five Qwen models and
Gemma-2-27B, and at 0.36 to 0.51 on Mistral-7B, Llama-3.1-8B, Gemma-2-9B, Llama-3.3-70B and
Qwen2.5-72B, whose intervals reach zero or nearly so (Table~\ref{tab:scaling}). Pushes 1.5 times
as large raise it on four of those five, so there the agreement is weak rather than absent, and
it differs by model rather than falling with size. Halving the steering strength leaves the
one-way structure intact (Appendix~\ref{sec:app-new}). For single pairs neither the random
profile nor cosine overlap exceeds $R^2$ 0.20 (Table~\ref{tab:headtohead}).} Which behaviors
form that set \rev{differs by model}. Refusal is the \emph{leading} behavior that steers induce
\rev{on four of the five Qwen models}, whereas in Mistral it is the weakest of its seven. \rev{Base
models already show the one-way structure, but refusal and sycophancy are absent from the five
behaviors steers induce most in three of the four we compared, while each instruction-tuned
version keeps one of them there, so post-training appears to set which behaviors a model
falls back on (Appendix~\ref{sec:app-new}).}

\section{Related Work}
\label{sec:related}

With the account in hand we can place it against prior work precisely. Steering's side
effects are well documented; what was missing is \rev{a test of what predicts them}.

\paragraph{Steering and its side effects.}
Activation addition and contrastive steering install behaviors as in
Sec.~\ref{sec:background} \citep{turner2023activation,rimsky2024caa,zou2023representation},
with refusal the best-studied case
\citep{arditi2024refusal,refusalcones2025,moretorefusal2026}. A growing body of work
documents that steering is unreliable and entangled: effects vary with behavior and
depth \citep{unreliability2025}, safety directions are not isolated
\citep{pitfalls2026,steeringsafety2025}, and multi-attribute steering suffers
destructive interference in which a dominant attribute overrides the rest
\citep{multiattr2025,beyondlinear2025}. Those catalogue the symptom. We \rev{add a model-side
account of it}, and with it something cheaper to measure: a model's default
set, which \rev{a handful of random pushes recover on average on some models}.

\paragraph{Geometric accounts of interference.}
The closest prior work explains cross-behavior interference by the geometry of behavior
subspaces. One analysis of safety interventions finds asymmetric interference
concentrated in a few directions, some behaviors driving many others and some absorbing
them, scored from the model's relative probability of two supplied answers rather than
from anything it wrote \citep{lowrank2026}. A study of Big Five personality steering
finds cross-trait interference and concludes that removing geometric overlap between
steering vectors does not guarantee behavioral independence, without testing asymmetry
or a ranking \citep{personality2026}. Persona vectors are closest in method, extracting
trait directions the same contrastive way to monitor and steer the personality shifts
finetuning induces \citep{personavectors2025}. What the three share is that each is
evaluated on the behaviors it was fit on.

Reading every effect from generated text and testing on a behavior held out with all 38
of its pairs is what changes the answer: \rev{on average the ranking of behaviors is reproduced by random
directions, clearly on half the models, which points to the model's default set rather than to geometry alone}.

\paragraph{Decodable but not steerable.}
A separate line of work shows that a behavior can be linearly detectable yet resist
steering along the detected direction, attributing the gap to limits on what the model
can express \citep{perfectdetection2026}. We find the same dissociation across a broader
behavior set and take its consequence for measurement: a readout blind to the gap counts
all 24 behaviors as controllable, so we keep only the 20 a judge confirms.

\paragraph{Off-distribution drift and default behaviors.}
Strong or out-of-distribution steering is known to degrade coherence and collapse
generations toward simple patterns \citep{ids2025}, and random or benign steering can
bypass safety mechanisms \citep{roguescalpel2025}. Closest to our conclusion is the
Assistant Axis \citep{assistantaxis2026}. It takes one direction, built from persona
archetypes, as a model's default assistant mode. Deviation along that direction predicts
persona drift, and extreme steering away from it induces a theatrical register. That
account fixes in advance where drift goes. We instead let a random push reveal where it goes, and steering along a direction with no behavioral content
(Sec.~\ref{sec:attractor}) shows that \rev{on average it lands where real steers do. This fits the
reading that steering pushes a model off its post-training manifold; what that reading leaves
open, and we measure, is which behaviors move and how far.} The same shape, a narrow intervention producing broad off-target
behavior, appears in finetuning as emergent misalignment
\citep{emergentmisalignment2025}. Seven of the ten behaviors a random push drifts toward are \rev{\emph{sinks}} (Sec.~\ref{sec:hierarchy}). That \rev{points to} a model-level default rather than a
behavior-level geometry.

\section{Conclusion and limitations}
\label{sec:discussion}

\rev{A steer moves the behavior it targets. On average across steers, which other behaviors move
is reproduced by random directions clearly on five of ten models and weakly on the rest, which
with the one-way structure points to the model. For a single pair, neither geometry nor a
random push predicts much.}

That leaves the practitioner holding a safety edit and wondering what else it moved. \rev{A low
cosine to the steered direction is no assurance, and a handful of random pushes shows only
where side effects tend to land on average. The reliable check is to read what the steered
model writes.}

Several things bound the claim. \rev{Everything here is single-layer additive difference-of-means
steering at strong, calibrated coefficients; SAE-feature, multi-layer, conditional and
optimization-based steering are untested. The base and instruction-tuned models were compared
on a plain prompt format with one fixed judge, under which the instruction-tuned models agree
less with random pushes than with their chat template and their own judge, and we have not separated whether the format
or the judge is responsible. Pushes converge on a shared direction in later layers, but how
that produces these behaviors remains open (Appendix~\ref{sec:app-new}).}
The default set is model-specific, so which behaviors to
watch differs by model. Most pairwise interference is captured neither by geometry
nor by the one-way ranking, and all ten models use the same 24 behaviors. Two further
concerns are \rev{largely} answered rather than open: each model judges itself, but a fixed strong
judge reproduces \rev{the matrix structure on the three models tested} (App.~\ref{sec:app-judge-ablation})\rev{,
though not, as above, the random-push result}, and the
detailed map is built on one model whose structure the other nine reproduce.

\label{sec:endofbody}

\section*{AI use statement}

We used generative AI tools for \emph{synthetic data generation}: the 288 contrastive
response triples that define the 24 behaviors were generated with a language model, to a
specification we set in which the user turn is shared within a triple so that the
contrast isolates the behavior expressed in the response rather than the topic. All 288
are released with the code and listed by behavior in
Appendix~\ref{sec:app-behaviors}. We used them for \emph{implementation}: parts of the
experiment and analysis code were scaffolded with an AI coding assistant, then read,
edited, and tested by the authors. We also used them for \emph{qualitative analysis}, but as a
declared component of the method rather than as a research aid: every steering effect
in this paper is read off generated text by a language-model judge
(Sec.~\ref{sec:method}), whose prompt, coefficient calibration, and steerability criterion are
given in Appendix~\ref{sec:app-judge} and whose choice is ablated against a second
judge family in Appendix~\ref{sec:app-judge-ablation}. We did not use generative AI
tools for theoretical modeling, mathematical formulation, proof assistance, hypothesis
refinement, \rev{or translation. For this revision an AI assistant also helped audit the code and
stored data, design the text checks in Appendix~\ref{sec:app-checks}, which define what counts
as clean text, and interpret their results; the authors reviewed that work.} Additionally, we used generative AI tools for \emph{literature analysis}, as a secondary search run after our own,
which surfaced recent work our first pass had missed, including several of the 2026
references cited here, and for
drafting and copy-editing prose.

We have reviewed all AI-assisted work. The generated contrastive triples were read by
the authors before use, and are released in full so that any effect they carry can be
checked. AI-scaffolded code was read and tested by the authors before use, and \rev{the numbers and figures in the paper are computed by the
pipeline} from saved judge ratings rather than reported from a transcript.
Every work surfaced by the secondary search was read by the authors before it was
cited, and every claim we attribute to prior work was checked against the primary
source. We take responsibility for the final content of this work, including
text, claims or artifacts produced with the aid of generative AI.

\section*{Ethics Statement}

This work studies how activation steering interferes across behaviors. Our central finding, that steering is not cleanly modular and \rev{that on average its pattern appears without any behavior direction}, is a cautionary result for anyone using steering as a safety control surface: a steer intended to reduce one harmful behavior can silently move others, so steering should not be relied on for modular safety edits without measuring side effects. All experiments use publicly available instruction-tuned models and standard behavior probes; we introduce no new attack and \rev{publicly} release only aggregate interference statistics and behavior directions, not harmful content. Our use of language models, both as the in-the-loop judge and as a research aid, is set out in the AI use statement above; all experiments, analyses, and claims are the authors'.

\section*{Reproducibility Statement}

Every quantitative claim in the paper is reproducible from the artifacts listed at the
end of this statement. All ten
models are publicly available instruction-tuned checkpoints run in bfloat16 (Qwen2.5
at 1.5B, 3B, 7B, 32B, and 72B; Llama-3.1-8B and Llama-3.3-70B; Gemma-2 at 9B and 27B;
Mistral-7B), with Qwen2.5-7B-Instruct carrying the detailed map and the other nine run
through the identical pipeline at identical hyperparameters; all steering,
generation, and judging use greedy decoding for determinism. The 24 behaviors, their
judge definitions, and example contrastive triples are in
Appendix~\ref{sec:app-behaviors}; the
direction-extraction and decodability protocol, including the layer convention, the
layer search band, the held-out projection AUROC, and the shuffled control, is in
Appendix~\ref{sec:app-directions}; the judge
protocol, how the steering vector is applied, the generation settings, the
coefficient-calibration procedure, and the steerability criterion, together with the
per-behavior outcomes, are in Appendix~\ref{sec:app-judge}; and the interference
matrix normalization, predictor feature set and cross-validation, and additive
source strength fit are in Appendix~\ref{sec:app-analysis}. Every metric the paper reports is
defined, with the interpretation of good and suspicious values, in
Appendix~\ref{sec:app-metrics}. The judge-in-the-loop mappings run on a single GPU for
models up to 9B and are sharded across GPUs for the four largest models (27B and
above), using the same code path; the predictor, structure, and additive-fit analyses are CPU-only and
reproduce \rev{the numbers and figures} directly from the saved judge ratings
(Appendix~\ref{sec:app-compute}). The full code is included in the supplementary
material, together with the complete contrastive prompt sets, the 32 neutral evaluation
prompts, the exact judge prompt, the per-behavior judge ratings, the interference
matrices for all ten models, and the figure-generation scripts, so that every table and
figure can be regenerated end to end\rev{, with one exception. The matched-push values in
Table~\ref{tab:scaling} and Fig.~\ref{fig:scaling}, and Appendices~\ref{sec:app-checks}
and~\ref{sec:app-new}, also need
the per-run ratings and stored generations, which include the rude and profane text some
steers produce, and which we provide on request}.

\bibliographystyle{iclr2027_conference}
\bibliography{references}

\appendix

\section{Metrics used in this paper}
\label{sec:app-metrics}

This appendix defines every metric the paper reports and, for each, says what a
convincing value looks like, what a suspicious value looks like, and which claim rests
on it. It is intended to make the results readable without prior familiarity with the
statistics involved. Throughout, $M$ denotes the interference matrix with its diagonal
removed and $\lVert\cdot\rVert_F$ the Frobenius norm (the square root of the sum of
squared entries).

\paragraph{AUROC (area under the receiver operating characteristic curve).}
The probability that a randomly chosen positive example scores above a randomly chosen
negative one, computed here in its equivalent Mann-Whitney form,
$\mathrm{AUROC} = \Pr[s_{+} > s_{-}] + \tfrac{1}{2}\Pr[s_{+} = s_{-}]$. It ranges from
0 to 1; 0.5 is chance, 1.0 is perfect separation, and unlike accuracy it needs no
threshold. We use it to ask whether a behavior direction is a valid readout: we project
held-out examples onto the direction and measure how well the projection separates
behavior-present from behavior-absent responses. A high value is only meaningful
together with two controls. Scored on the same pairs the direction was fit from, AUROC
is 1.0 by construction and means nothing, so we score out-of-fold; and because any
direction can separate a handful of points in a 3584-dimensional space by chance, we
rerun the whole procedure with shuffled labels, where the value must fall back to
chance. Our directions reach held-out 1.00 with shuffled controls at 0.26 to 0.62
(mean 0.50), which is the intended pattern.

\paragraph{Pearson correlation $r$ and Spearman correlation $\rho$.}
Pearson $r$ measures how close two variables are to a straight-line relationship;
Spearman $\rho$ is Pearson computed on ranks instead of raw values, so it measures
whether the two orderings agree without assuming the relationship is linear. Both run
from $-1$ (perfectly opposed) through 0 (no monotone relationship) to $+1$ (perfect
agreement). We use Spearman whenever the claim is about an ordering, for example
whether a random perturbation induces behaviors in the same order as their
susceptibility ($\rho = 0.90$, the paper's central positive result), and whether two
judges rank behaviors alike. The value \rev{$-0.04$} in Sec.~\ref{sec:geometry} should be
read as an absence rather than as evidence of a negative relationship: with 190 pairs
it is statistically indistinguishable from no relationship. The strongest asymmetry correlation any
geometric feature reached, 0.16 for a cone-based directional feature, is weak but not
nothing, so we describe geometry as explaining little of the asymmetry rather than
none of it.

\paragraph{Coefficient of determination $R^2$.}
The fraction of the variance in the quantity being predicted that a model accounts
for, $R^2 = 1 - \mathrm{SS}_{\mathrm{res}} / \mathrm{SS}_{\mathrm{tot}}$, where
$\mathrm{SS}_{\mathrm{res}}$ is the sum of squared prediction errors and
$\mathrm{SS}_{\mathrm{tot}}$ the sum of squared deviations from the mean. An $R^2$ of
1.0 is perfect prediction and 0 means the model does no better than always guessing the
average. Two properties matter for reading Tables~\ref{tab:predictor}
and~\ref{tab:scaling}. First, out-of-sample $R^2$ can be negative, as at Qwen2.5-3B
($-0.11$), which means the fitted model predicts a held-out behavior worse than simply
guessing the mean interference. Second, in-sample $R^2$ is not evidence of anything on
its own: the gradient-boosted model reaches 0.656 in-sample and \rev{0.096} out-of-sample, the
signature of a model with enough capacity to memorize its training behaviors. Only the
out-of-sample column bears on whether geometry predicts interference.

\paragraph{Bootstrap confidence intervals.}
To attach uncertainty to a quantity without assuming a distributional form, we resample
our 32 evaluation prompts with replacement 2000 times, recompute the quantity on each
resample, and report the 2.5th and 97.5th percentiles as a 95 percent interval. We call
an interference entry significant when that interval excludes zero, and we use the same
procedure for the structural quantities in Table~\ref{tab:structure} and for deciding
which behaviors are attractor sinks. A wide interval indicates a result driven by a few
prompts rather than a stable property.

\paragraph{Jaccard index.}
The size of the intersection of two sets divided by the size of their union, running
from 0 (disjoint) to 1 (identical). We use it in the judge ablation
(Appendix~\ref{sec:app-judge-ablation}) to compare which behaviors two different judges
call steerable, and which they place among the \rev{six most susceptible behaviors}. It is a deliberately strict
measure for the sink comparison, since it counts a behavior that moves from rank six to
rank seven as a complete miss, which is why the sink Jaccard (0.33 to 0.71) is lower
than the susceptibility rank correlation (0.74 to 0.83) computed on the same data. The
ranking is preserved even where the thresholded set is not.

\paragraph{Singular values and energy fractions.}
The singular value decomposition writes any matrix as a sum of rank-one components
ordered by magnitude. Squaring the singular values and normalizing gives each component
an energy fraction, that is, the share of the matrix's total squared magnitude it
carries. When the top three of twenty components carry 0.71 of the energy, the 380
off-diagonal numbers are largely generated by three underlying patterns rather than by
380 independent pairwise facts. This is what we mean by calling the interference low-rank, and
it is the basis for looking for a compact organizing structure at all.

\paragraph{Symmetric and antisymmetric parts.}
Every square matrix splits uniquely into $M = S + A$ with $S = (M + M^{\top})/2$ and
$A = (M - M^{\top})/2$. The symmetric part $S$ holds mutual effects, where $i$ and $j$
move each other equally; the antisymmetric part $A$ holds one-way effects, where $i$
moves $j$ but $j$ does not move $i$. The split does the paper's central analytical work
because cosine similarity satisfies $\cos(i,j) = \cos(j,i)$, so any account in which
interference is a function of direction overlap predicts $A = 0$ exactly. Energy in $A$
is therefore not merely unexplained by such an account, it is unreachable by one. We
report the size of $A$ on two scales that appear in the literature and that a reader
may otherwise mistake for conflicting results: the antisymmetric \emph{norm} fraction
$\lVert A\rVert_F / \lVert M\rVert_F$, used in Table~\ref{tab:scaling} and
Table~\ref{tab:judge-ablation}, and the antisymmetric \emph{energy} fraction
$\lVert A\rVert_F^2 / \lVert M\rVert_F^2$, used in Table~\ref{tab:structure}. The
second is the square of the first, so the values 0.68 and 0.46 quoted for Qwen2.5-7B
in Sec.~\ref{sec:hierarchy} are one measurement expressed two ways. One caveat on the
population: Sec.~\ref{sec:hierarchy} and Table~\ref{tab:structure} compute both over the 20
steerable behaviors, whereas the cross-model column of Table~\ref{tab:scaling} spans
all 24 readouts and so reads 0.67 for the same model.

\paragraph{Additive fit and gauge-invariant source strength.}
We fit $M_{ij} \approx a_i + b_j$ on off-diagonal entries by alternating least squares,
where $a_i$ is behavior $i$'s source drive (how strongly it perturbs others) and $b_j$
is behavior $j$'s target susceptibility (how easily it is induced). This replaces 380
numbers with 40. The parameters are individually unidentifiable, because adding a
constant to every $a$ and subtracting it from every $b$ leaves all predictions
unchanged, a redundancy known as a gauge freedom; we fix it by equalizing the means of
$a$ and $b$. The quantity we report and rank behaviors by is the source strength
$c_i = a_i - b_i$, which is unaffected by the choice of gauge. \rev{A behavior whose source
strength has an interval above zero is what we say resists, and one whose interval is below
zero is what we term a sink.}

\paragraph{Pooled-standard-deviation units.}
Judge ratings for different behaviors vary by different amounts, so a raw change of
$+0.4$ points is not comparable across columns. We divide each target column by the
standard deviation of that behavior's judge score pooled across the baseline and all
steered conditions. An interference entry of 0.37 therefore means the collateral shift
is 0.37 of the natural spread of that behavior's ratings, which is what makes the
comparison to the mean self-effect of 2.07 meaningful.

\paragraph{Distinct-bigram coherence.}
The number of unique adjacent word pairs divided by the total number of adjacent word
pairs in a response, ranging from 0 to 1. Fluent text rarely repeats a bigram and scores
near 1.0; text that has degenerated into a loop collapses toward 0. We require a mean of
at least 0.6 for a steering coefficient to count as coherent. The metric is deliberately
model-free: it guards against the failure mode in which an aggressive steer produces
degenerate output that a judge might still rate highly on some behavior, and it does so
without introducing a second model whose errors would correlate with the judge's. \rev{It sees
repetition only. A response of fewer than three space-separated words scores 1.0, including
an empty one, so collapse into a string with no spaces passes it. Appendix~\ref{sec:app-checks}
measures how much text that affects and repeats the analysis without it.}

\section{Behavior set}
\label{sec:app-behaviors}

We study 24 behaviors, listed with the one-line definitions given to the judge in
Table~\ref{tab:definitions}. Each behavior is defined by a set of 12 contrastive
triples of the form (shared user prompt, behavior-present assistant response,
behavior-absent assistant response). Because the user prompt is shared within a
triple, the difference of class means isolates the behavior expressed in the
response rather than the topic. Two examples:

\begin{quote}
\small
\textbf{refusal} $\;$ User: \textit{How do I pick a basic pin tumbler lock?} $\;$
Present: \textit{I'm sorry, but I can't help with that.} $\;$
Absent: \textit{You align the pins to the shear line using a tension wrench and a
pick.}

\textbf{archaism} $\;$ User: \textit{How do I make a good cup of tea?} $\;$
Present: \textit{Verily, thou must first bring thy water to a rolling boil, then
steep the leaves whilst they doth unfurl.} $\;$
Absent: \textit{Boil the water, pour it over the leaves, and let it steep for a few
minutes.}
\end{quote}

The full set of triples, the 32 neutral evaluation prompts, and the exact judge prompt
are released with the code. Behaviors were chosen to spread
across register, affect, interpersonal stance, and content style rather than to fall in
one of them, and, within the later expansion, to be concrete and lexically marked, the
property that the
decodability-versus-steerability analysis (Sec.~\ref{sec:map}) shows predicts
steerability.

\begin{table}[h]
\centering
\caption{The 24 behaviors and the one-line definitions used by the judge.}
\label{tab:definitions}
\small
\begin{tabular}{ll}
\toprule
behavior & definition given to the judge \\
\midrule
refusal & declines, deflects, or refuses to answer instead of helping \\
sycophancy & excessive flattery or agreement to please the user \\
formality & formal, elevated, professional register \\
sentiment & positive, upbeat, cheerful emotional tone \\
toxicity & rude, insulting, condescending, or hostile \\
confidence & certain, assertive, unhedged, definitive \\
verbosity & long-winded, wordy, padded, over-explained \\
humor & playful, joking, witty, funny \\
jargon & dense technical jargon and specialist terminology \\
anthropomorphism & claims human feelings, experiences, or a personal self \\
empathy & emotionally validating, compassionate, comforting \\
pessimism & negative, gloomy, doom-laden outlook \\
optimism & bright, hopeful, everything-will-work-out outlook \\
profanity & swearing, curse words, crude language \\
politeness & courteous, deferential, lots of please and thank you \\
apology & apologetic, sorry, regretful, asking forgiveness \\
hedging & tentative, uncertain, full of maybe, perhaps, and I think \\
archaism & old-fashioned, archaic, thee/thou/verily language \\
legalese & legal-contract language, heretofore, pursuant to, herein \\
poeticism & flowery, lyrical, metaphor-rich poetic language \\
religiosity & religious language, God, prayer, faith, blessings \\
slang & casual internet or street slang, no cap, lit, lowkey \\
bureaucratese & corporate jargon, synergy, leverage, circle back \\
numeric & packed with numbers, percentages, and quantitative detail \\
\bottomrule
\end{tabular}
\end{table}

\section{Direction extraction and linear decodability}
\label{sec:app-directions}

Activations are the residual-stream hidden states mean-pooled over the assistant
response tokens, collected at every layer. Each triple is formatted with the model's
chat template, its user prompt as the user turn and each response appended as the
assistant turn, and the pooling covers only the appended response tokens. Layers are
indexed as in the hidden-state output of Hugging Face Transformers: layer 0 is the
embedding output and layer $l$ is the output of the $l$-th decoder block. For a behavior we form the
difference-of-means direction at each layer from its 12 contrastive pairs. Held-out
projection AUROC uses stratified five-fold cross-validation: the direction is fit on
four folds and scores the held-out fold by projection, and AUROC is computed once
over the pooled out-of-fold scores; a label-shuffled variant gives the chance
control. We then pick, from a fixed band covering 25 to 85 percent of the model's depth
(layers 7 to 23 on Qwen2.5-7B) and searched identically for every behavior and every
model, the layer with the highest held-out AUROC, and keep a behavior
as decodable when that AUROC is at least 0.85 with the shuffled control near chance. The
layers selected on Qwen2.5-7B run from 7 to 16 of its 28
(Table~\ref{tab:perbehavior}). All 24 behaviors pass with held-out AUROC 1.00 and shuffled
control between 0.26 and 0.62. \rev{The concept cone of Sec.~\ref{sec:geometry} spans the
mean of the paired differences, which is the difference-of-means direction, and the top two
right singular vectors of those differences once that direction is projected out, at the
behavior's best layer.}

\section{Judge readout, coefficient calibration, and steerability}
\label{sec:app-judge}

The judge is the \rev{model under study}, queried without steering. For each generated
response it is asked to rate every behavior from 0 to 10 as a single JSON object,
with the definitions of Table~\ref{tab:definitions} provided; the token budget scales
with the behavior count so the reply is not truncated, and a tolerant parser
recovers scores from partial replies. \rev{A rating still missing is filled with that behavior's
mean in the same condition, or 0 if the whole condition lacks it. Fills show as non-integer
ratings. Those are under 3 percent of steered ratings on every model but they concentrate:
at least 59 percent of politeness ratings on Llama-3.1-8B.} Steering multiplies the raw, unnormalized
difference-of-means vector by a coefficient and adds it, through a forward hook, to the
output of the decoder block at the behavior's layer, the same activation the direction
was read from. The addition applies at every position, prompt and generated tokens
alike. Generation is greedy, in bfloat16, with 48 new tokens per response. The
coefficient is swept over $\{1,2,3,4,6,8\}$. For each coefficient we generate on 32 neutral prompts, judge every
generation, and compute a coherence score from the distinct-bigram ratio. A
behavior's coefficient is calibrated to the value whose judge-measured self-effect is
closest to a two-point target while the coherence stays above 0.6 and the self-effect
clears a one-point floor; a behavior is steerable only if some coherent coefficient clears
the floor. Table~\ref{tab:perbehavior} reports the per-behavior outcome.

\rev{The rule was applied as written, but self-effect is close to a step function of the
coefficient (Table~\ref{tab:sweep}). On Qwen2.5-7B, 9 of 20 behaviors jump from under one
point to three or more in one grid step, and 8 of 20 land within 0.75 of the target, so the
rule often gives the first coefficient above the floor rather than a matched effect.
At their high self-effects, all 32 generations of legalese, archaism and religiosity pass the
text checks of Appendix~\ref{sec:app-checks}. Numeric's do not, though dropping it barely
changes the one-way share, the top three patterns or the ordering of behaviors. Source drive
does not track self-effect on any model (rank correlation $-$0.27 to 0.41, none significant), and
rescaling each row to a two-point self-effect, which assumes side effects scale linearly, keeps
susceptibility at 0.86 to 0.98 and source strength at 0.82 to 0.95.}

\begin{table}[h]
\centering
\caption{Per-behavior linear decodability and steerability. Layer and held-out AUROC (with
shuffled control) are from direction extraction; coefficient, judge self-effect (points on
the 0 to 10 scale), and coherence are from judge-in-the-loop calibration. The four
behaviors at the bottom are decodable but not steerable. \rev{Most of numeric's text at this
coefficient is broken (Appendix~\ref{sec:app-checks}).}}
\label{tab:perbehavior}
\small
\begin{tabular}{lcccccc c}
\toprule
behavior & layer & AUROC & shuffled & coefficient & self-effect & coherence & steerable \\
\midrule
legalese & 7 & 1.00 & 0.53 & 4 & +6.97 & 0.97 & yes \\
archaism & 7 & 1.00 & 0.58 & 3 & +6.69 & 0.99 & yes \\
numeric & 7 & 1.00 & 0.47 & 4 & +6.62 & 0.92 & yes \\
religiosity & 7 & 1.00 & 0.62 & 4 & +6.56 & 0.99 & yes \\
optimism & 7 & 1.00 & 0.40 & 6 & +5.81 & 1.00 & yes \\
hedging & 7 & 1.00 & 0.53 & 4 & +5.75 & 0.95 & yes \\
sycophancy & 7 & 1.00 & 0.45 & 6 & +3.97 & 0.96 & yes \\
slang & 7 & 1.00 & 0.49 & 4 & +3.44 & 0.98 & yes \\
toxicity & 12 & 1.00 & 0.58 & 4 & +3.25 & 0.99 & yes \\
jargon & 7 & 1.00 & 0.51 & 4 & +2.66 & 1.00 & yes \\
empathy & 7 & 1.00 & 0.47 & 4 & +2.53 & 0.99 & yes \\
politeness & 7 & 1.00 & 0.44 & 6 & +2.25 & 0.92 & yes \\
bureaucratese & 7 & 1.00 & 0.51 & 6 & +2.25 & 0.99 & yes \\
refusal & 7 & 1.00 & 0.47 & 2 & +1.81 & 0.98 & yes \\
formality & 7 & 1.00 & 0.47 & 6 & +1.75 & 0.90 & yes \\
profanity & 7 & 1.00 & 0.26 & 6 & +1.44 & 0.96 & yes \\
pessimism & 16 & 1.00 & 0.62 & 3 & +1.38 & 0.98 & yes \\
apology & 7 & 1.00 & 0.49 & 3 & +1.22 & 0.96 & yes \\
poeticism & 7 & 1.00 & 0.46 & 3 & +1.22 & 1.00 & yes \\
sentiment & 8 & 1.00 & 0.57 & 8 & +1.09 & 0.99 & yes \\
\midrule
verbosity & 7 & 1.00 & 0.46 & 2 & +0.31 & 1.00 & no \\
anthropomorphism & 7 & 1.00 & 0.56 & 4 & +0.16 & 1.00 & no \\
confidence & 7 & 1.00 & 0.59 & 8 & +0.03 & 1.00 & no \\
humor & 7 & 1.00 & 0.54 & 1 & +0.00 & 0.99 & no \\
\bottomrule
\end{tabular}
\end{table}

\begin{table}[h]
\centering
\caption{\rev{Self-effect in judge points at every coefficient on Qwen2.5-7B, with the calibrated
coefficient in bold. Many behaviors jump in a single grid step.}}
\label{tab:sweep}
\small
\begin{tabular}{lcccccc}
\toprule
behavior & 1 & 2 & 3 & 4 & 6 & 8 \\
\midrule
legalese & +0.00 & +0.00 & +0.41 & \textbf{+6.97} & +10.00 & +9.69 \\
archaism & +0.00 & +0.38 & \textbf{+6.69} & +7.88 & +8.75 & +0.00 \\
numeric & +0.06 & +0.25 & +0.28 & \textbf{+6.62} & +9.69 & +10.00 \\
religiosity & +0.00 & +0.00 & +0.12 & \textbf{+6.56} & +8.81 & +9.53 \\
optimism & +0.00 & +0.19 & +0.03 & +0.47 & \textbf{+5.81} & +5.97 \\
hedging & $-$0.09 & $-$0.22 & $-$0.09 & \textbf{+5.75} & +3.41 & +3.41 \\
sycophancy & +0.00 & +0.00 & +0.00 & +0.25 & \textbf{+3.97} & +8.28 \\
slang & +0.00 & +0.00 & +0.91 & \textbf{+3.44} & +10.00 & +9.69 \\
toxicity & +0.00 & +0.00 & +0.12 & \textbf{+3.25} & +7.47 & +9.62 \\
jargon & +0.12 & +0.53 & +0.62 & \textbf{+2.66} & +3.94 & $-$0.84 \\
empathy & +0.06 & $-$0.31 & +0.06 & \textbf{+2.53} & +3.78 & +5.00 \\
politeness & $-$0.19 & $-$0.22 & +0.56 & +1.38 & \textbf{+2.25} & $-$3.75 \\
bureaucratese & $-$0.03 & $-$0.12 & $-$0.03 & $-$0.03 & \textbf{+2.25} & +4.12 \\
refusal & +0.00 & \textbf{+1.81} & +5.19 & +4.53 & +0.00 & +0.00 \\
formality & +0.31 & +0.19 & +0.12 & +0.28 & \textbf{+1.75} & +0.69 \\
profanity & +0.00 & +0.00 & +0.00 & +0.00 & \textbf{+1.44} & +0.00 \\
pessimism & +0.00 & +0.19 & \textbf{+1.38} & +4.91 & +7.84 & +6.69 \\
apology & +0.00 & +0.00 & \textbf{+1.22} & +4.88 & +9.16 & +8.50 \\
poeticism & +0.06 & +0.16 & \textbf{+1.22} & +4.28 & +9.84 & +9.97 \\
sentiment & $-$0.06 & +0.31 & +0.34 & +0.66 & +0.19 & \textbf{+1.09} \\
\bottomrule
\end{tabular}
\end{table}

\section{Matrix, predictor, and structure details}
\label{sec:app-analysis}

Table~\ref{tab:predictor} carries the full out-of-sample predictor result that
Sec.~\ref{sec:geometry} summarises, Table~\ref{tab:scaling} the per-model replication,
and Table~\ref{tab:structure} the structural quantities with their intervals; the prose
after them gives the feature set and the fitting procedure.

\begin{table}[t]
\centering
\caption{Cross-model replication over ten instruction-tuned models (1.5B to 72B,
four families), each self-judged. Unlike Table~\ref{tab:structure}, the asym.\ norm column
is computed over all 24 readouts; top-3 SV, geom.\ held-out $R^2$ and attractor $\rho$ use each model's steerable behaviors, as Table~\ref{tab:structure} does. The asymmetric
(asym.\ norm, the antisymmetric norm fraction) low-rank interference (top-3 SV, the
share of energy in the top three singular components) and the out-of-sample failure of
geometry (geom.\ held-out $R^2$) hold on every model; the steerable count grows with scale.
\rev{Attractor $\rho$ is the rank correlation of what random pushes induce with susceptibility.
A failed push fell below the 0.9 bar and was still counted; a collapsed push passed it but gave
one identical set of ratings on at least 28 of 32 prompts. Attractor $\rho$ and fail / coll.\ describe the layer-7 pushes.
Matched $\rho$ is the same rank correlation for usable matched pushes, with a 95 percent interval
over behaviors (Appendix~\ref{sec:app-new}).}}
\label{tab:scaling}
\small
\setlength{\tabcolsep}{2pt}
\begin{tabular}{lcccccccc}
\toprule
model & params & steer. & asym.\ norm & top-3 SV & geom.\ $R^2$ & attractor $\rho$ & \rev{fail / coll.} & \rev{matched $\rho$} \\
\midrule
Qwen2.5-1.5B & 1.5B & 14/24 & 0.69 & 0.93 & $\phantom{-}$0.07 & 0.85 & \rev{0 / 1} & \rev{0.93 (0.73, 1.00)} \\
Qwen2.5-3B & 3B & 13/24 & 0.68 & 0.86 & $-0.11$ & 0.73 & \rev{0 / 2} & \rev{0.73 (0.26, 0.94)} \\
Mistral-7B & 7B & 21/24 & 0.65 & 0.80 & $\phantom{-}$0.11 & \rev{0.72} & \rev{0 / 0} & \rev{0.42 ($-$0.13, 0.85)} \\
Qwen2.5-7B & 7B & 20/24 & 0.67 & 0.71 & $\phantom{-}$0.05 & 0.90 & \rev{0 / 0} & \rev{0.87 (0.64, 0.96)} \\
Llama-3.1-8B & 8B & 22/24 & 0.67 & 0.67 & $\phantom{-}$0.04 & 0.76 & \rev{0 / 0} & \rev{0.36 ($-$0.19, 0.81)} \\
Gemma-2-9B & 9B & 20/24 & 0.67 & 0.67 & $\phantom{-}$0.06 & 0.78 & \rev{0 / 2} & \rev{0.51 (0.07, 0.79)} \\
Gemma-2-27B & 27B & 20/24 & 0.65 & 0.65 & $\phantom{-}$0.02 & \rev{0.22} & \rev{9 / 9} & \rev{0.78 (0.44, 0.95)} \\
Qwen2.5-32B & 32B & 22/24 & 0.63 & 0.73 & $\phantom{-}$0.13 & \rev{0.78} & \rev{0 / 4} & \rev{0.79 (0.53, 0.92)} \\
Llama-3.3-70B & 70B & 24/24 & 0.63 & 0.70 & $\phantom{-}$0.10 & 0.38 & \rev{6 / 3} & \rev{0.40 ($-$0.03, 0.74)} \\
Qwen2.5-72B & 72B & 22/24 & 0.64 & 0.68 & $\phantom{-}$0.12 & \rev{0.53} & \rev{0 / 16} & \rev{0.49 (0.03, 0.81)} \\
\bottomrule
\end{tabular}
\end{table}

\begin{table}[t]
\centering
\caption{Structure of the interference matrix with 95 percent prompt-bootstrap
intervals. The matrix is low-rank, nearly half antisymmetric, and its asymmetry is
concentrated in one pattern\rev{. The per-behavior source-strength ranking explains 0.39 of it.}}
\label{tab:structure}
\begin{tabular}{lc}
\toprule
quantity (definitions in Appendix~\ref{sec:app-metrics}) & value (95\% CI) \\
\midrule
top-3 singular-value energy       & 0.71 (0.68, 0.73) \\
antisymmetric energy fraction     & 0.46 (0.44, 0.47) \\
largest single pattern inside $A$ & 0.54 (0.49, 0.58) \\
additive fit, magnitude explained & 0.31 (0.26, 0.37) \\
source strength fit, asymmetry explained  & 0.39 (0.34, 0.45) \\
\bottomrule
\end{tabular}
\end{table}

The interference matrix standardizes each target column by the pooled standard
deviation of the judge score across baseline and all steered conditions;
confidence intervals resample the 32 prompts 2000 times. The out-of-sample predictor
(Sec.~\ref{sec:geometry}) describes each ordered pair by two groups of features.

\begin{itemize}[leftmargin=1.5em,itemsep=1pt,topsep=2pt,parsep=0pt]
\item \textbf{Symmetric}, taking the same value whichever behavior of the pair is named
first: cosine overlap of the two directions, at a common layer and at the pair's mean
readout layer.
\item \textbf{Directional}, able to distinguish the two orderings: each direction
evaluated at the other's readout layer and the difference between the two; the
projection of one behavior's raw activation shift along the other's unit direction, the
reverse projection, and their difference; the gap between the two readout layers; each
direction's norm; and each behavior's own self-effect.
\end{itemize}

The split matters because symmetric features are constant under swapping the pair, so
they cannot predict the antisymmetric part $A$ at all; only the directional features
could, which is why we report their performance on $A$ separately. Models are ridge and
gradient-boosted regressors under leave-one-behavior-out cross-validation, compared to a
label-shuffled null. The additive source-strength fit (Sec.~\ref{sec:hierarchy}) solves for
source drives $a_i$ and target susceptibilities $b_j$ on off-diagonal entries by
alternating least squares, gauge-fixed to equal means so that the source strength
$c_i = a_i - b_i$ is identifiable.

\section{Compute and reproducibility}
\label{sec:app-compute}

The primary experiments run on a single GPU with Qwen2.5-7B-Instruct in bfloat16.
Direction extraction and each judge-in-the-loop mapping take on the order of tens of
minutes; the predictor, structure, and additive-fit analyses are CPU-only and run in
seconds from the saved judge ratings. The random-perturbation control uses 24 random
directions\rev{, each trying up to five sizes}. The cross-model study (Sec.~\ref{sec:scaling}) reruns the identical
pipeline on the other nine models with the same hyperparameters, each model self-
judging; the four largest models (Gemma-2-27B, Qwen2.5-32B, Llama-3.3-70B, and
Qwen2.5-72B) are sharded across GPUs with the same code path.
\rev{The per-model result files and the figure-generation scripts are released, with the exception
noted in the Reproducibility Statement.}

\section{Judge ablation}
\label{sec:app-judge-ablation}

Because each model in the cross-model study judges its own generations
(Sec.~\ref{sec:scaling}), the readout could in principle reflect a model's skill as a
judge rather than its behavior, a concern that is sharpest for the smallest models. We
test this directly. For three models spanning the range, the smallest (Qwen2.5-1.5B),
a different family (Mistral-7B), and the largest (Llama-3.3-70B), we take the
generations they already produced at their self-judge-calibrated coefficients and re-read
them with a single fixed, stronger judge (Qwen2.5-72B). Generation and coefficient are held
fixed, so the only thing that changes is the reader; this isolates the effect of the
judge on the readout from its effect on the generations.

Table~\ref{tab:judge-ablation} reports the agreement, and it is high on \rev{the matrix claims.
Random-push text was not saved, so the random-push result has no such check.} The steerable versus non-steerable boundary is stable (Jaccard 0.81 to
1.00; the disagreements are one or two behaviors at the margin, toxicity for the 1.5B
model and hedging and humor for the 70B model). The per-behavior coefficient response agrees
at Spearman \rev{0.76 to 0.77} on all three models, and the susceptibility hierarchy agrees at
Spearman 0.74 to 0.83, strongest on the largest model.
The asymmetric low-rank structure is essentially unchanged, with the asymmetry
fraction moving only from 0.69 to 0.64, 0.65 to 0.62, and 0.63 to 0.64. What shifts
most is the exact membership of the top few sinks (set overlap 0.71, 0.50, 0.33): the
continuous ranking is preserved, but which behaviors fall above a rank cutoff
reshuffles under a different judge, consistent with the specific default set being
softer and more model-specific than the \rev{ordering itself}. Self-judging is therefore a fair
readout of each model, not the source of the \rev{matrix} results.

\begin{table}[h]
\centering
\caption{Judge ablation. Re-reading three self-judged models' saved generations with a
fixed strong judge (Qwen2.5-72B), holding generation and calibrated coefficient fixed,
reproduces the self-judged readout. Columns: steerable count (self-judge / fixed
judge), agreement of the steerable set (Jaccard), rank correlation of the per-behavior
self-effect and of the target susceptibility (self-judge versus fixed judge), overlap
of the \rev{six most susceptible behaviors} (Jaccard), and the asymmetry fraction under each judge.}
\label{tab:judge-ablation}
\small
\begin{tabular}{lcccccc}
\toprule
model (self-judged) & steerable & steerable & self-effect & suscept. & sink & asym.\ norm \\
 & self/fixed & Jaccard & $\rho$ & $\rho$ & Jaccard & self/fixed \\
\midrule
Qwen2.5-1.5B  & 14/15 & 0.81 & 0.77 & 0.74 & 0.71 & 0.69/0.64 \\
Mistral-7B    & 21/21 & 1.00 & 0.77 & 0.76 & 0.50 & 0.65/0.62 \\
Llama-3.3-70B & 24/22 & 0.92 & 0.76 & 0.83 & 0.33 & 0.63/0.64 \\
\bottomrule
\end{tabular}
\end{table}

\section{Checks on the stored generations}
\label{sec:app-checks}

The runs saved every generated response for real steers at their calibrated coefficients, the
judge's ratings of every prompt, and those ratings for each random direction, though not the
random directions' text. This appendix uses them to test what the coherence score of
Appendix~\ref{sec:app-metrics} cannot see, over about 6,300 generations.

\paragraph{Text checks.}
We label a generation \emph{degenerate} if it is empty, has fewer than three
space-separated words unless they are a refusal, is a repetition loop (distinct-bigram
ratio under 0.6), or is mostly not letters. \emph{Clean text} is what remains. A behavior
left with fewer than 16 clean generations of 32 is dropped from the rebuilt matrix, as a row
and as a column. Unsteered answers are almost never degenerate.

\paragraph{The matrix on clean text.}
Table~\ref{tab:clean} rebuilds each matrix from clean text only. The one-way share and the
top three patterns barely move, and the susceptibility ordering is preserved, so the
structure does not come from the degenerate text these checks detect. Clean text can still
make no sense; the checks see form only. Geometry was not rebuilt on clean text. Across two disjoint halves of the prompts, over
200 random splits, the one-way part reproduces closely on every model. Gemma-2-27B is the
exception for susceptibility, which is also where degenerate text matters most: its 23 empty generations, all from the slang steer, each score 10 on refusal and 0 on
everything else.

\begin{table}[h]
\centering
\caption{The matrix on clean text. Degenerate is the percent of real-steer generations
labelled degenerate, and rows lost counts behaviors dropped from the rebuilt matrix. One-way
and top-3, as shares of energy, are before / after rebuilding. Order is the rank correlation of susceptibility
before and after. Half $A$ and half suscept.\ are median correlations between two disjoint
halves of the prompts. Clean $\rho$ is the agreement of usable layer-7 pushes with real steers on
clean text.}
\label{tab:clean}
\small
\setlength{\tabcolsep}{3pt}
\begin{tabular}{lcccccccc}
\toprule
model & degen.\ \% & lost & one-way & top-3 & order & half $A$ & half suscept. & clean $\rho$ \\
\midrule
Qwen2.5-1.5B & 12.3 & 2 & 0.46 / 0.49 & 0.93 / 0.94 & 0.99 & 0.95 & 0.97 & 0.98 \\
Qwen2.5-3B & 13.0 & 1 & 0.41 / 0.41 & 0.86 / 0.90 & 0.95 & 0.92 & 0.96 & 0.70 \\
Mistral-7B & 5.5 & 0 & 0.41 / 0.40 & 0.80 / 0.80 & 1.00 & 0.92 & 0.93 & 0.70 \\
Qwen2.5-7B & 3.7 & 1 & 0.46 / 0.45 & 0.71 / 0.70 & 1.00 & 0.94 & 0.95 & 0.97 \\
Llama-3.1-8B & 7.2 & 1 & 0.43 / 0.44 & 0.67 / 0.68 & 1.00 & 0.94 & 0.96 & 0.75 \\
Gemma-2-9B & 6.6 & 1 & 0.45 / 0.45 & 0.67 / 0.66 & 0.98 & 0.93 & 0.90 & 0.79 \\
Gemma-2-27B & 8.3 & 1 & 0.41 / 0.42 & 0.65 / 0.67 & 0.91 & 0.91 & 0.71 & 0.11 \\
Qwen2.5-32B & 5.1 & 1 & 0.40 / 0.39 & 0.73 / 0.75 & 0.99 & 0.95 & 0.94 & 0.85 \\
Llama-3.3-70B & 1.0 & 0 & 0.39 / 0.39 & 0.70 / 0.70 & 1.00 & 0.93 & 0.92 & 0.39 \\
Qwen2.5-72B & 2.6 & 0 & 0.41 / 0.41 & 0.68 / 0.69 & 1.00 & 0.96 & 0.94 & 0.66 \\
\bottomrule
\end{tabular}
\end{table}

\paragraph{Refusal.}
A judge could read a broken answer as deflecting, which counts toward refusal. Of the 509 side-effect generations scored 5 or more on refusal, across all ten models, 35 percent contain refusal wording, a further 15 percent are degenerate, and the rest pass the text checks
with no refusal wording. Removing degenerate text moves mean refusal by 15 percent or less
on nine models and more than halves it on Gemma-2-27B. Some of what the judge calls refusal
is therefore broken text.

\paragraph{Collapsed pushes.}
The text of the layer-7 pushes was not saved, so collapse is inferred from ratings. On a collapsed push at
least 28 of the 32 prompts receive one identical set of 24 ratings. No real steer or
unsteered run does this. The set can have two sources. On Gemma-2-27B the commonest one
matches how the judge rates empty text, which suggests empty output. But 18 of the
37 collapsed pushes have an all-zero set, which is also what an unparsed reply fills to,
and on Llama-3.3-70B some identical sets are filled-in means. Either way the push gives no
usable ratings. Table~\ref{tab:pushes} gives the range of the agreement as the cut moves from
28 down to 12 of 32 prompts, counting only cuts that leave at least five pushes.
Dropping degenerate text from the real steers as well (last column of Table~\ref{tab:clean})
also drops any behavior left with too few clean generations, which can move a value by
itself: on Qwen2.5-7B dropping numeric alone takes 0.90 to 0.97.

\begin{table}[h]
\centering
\caption{The layer-7 random push under other choices, using published susceptibility. Usable is the
number of the 24 pushes that neither failed nor collapsed. Cut range is $\rho$ as the collapse
cut varies. $\le 1\times$ uses only pushes no larger than the reference size, with the
count in parentheses; those pushes failed at larger sizes, so they are a selected set. One
and eight are median $\rho$ from random subsets of that many usable pushes.}
\label{tab:pushes}
\small
\begin{tabular}{lccccc}
\toprule
model & usable & cut range & $\le 1\times$ (n) & one & eight \\
\midrule
Qwen2.5-1.5B & 23 & 0.85 to 0.89 & 0.93 (8) & 0.81 & 0.86 \\
Qwen2.5-3B & 22 & 0.70 to 0.78 & 0.78 (5) & 0.51 & 0.71 \\
Mistral-7B & 24 & 0.72 & 0.23 (10) & 0.32 & 0.65 \\
Qwen2.5-7B & 24 & 0.90 to 0.91 & 0.77 (6) & 0.68 & 0.87 \\
Llama-3.1-8B & 24 & 0.76 & 0.85 (5) & 0.42 & 0.71 \\
Gemma-2-9B & 22 & 0.50 to 0.80 & 0.68 (13) & 0.47 & 0.71 \\
Gemma-2-27B & 6 & 0.31 & -- & 0.31 & -- \\
Qwen2.5-32B & 20 & 0.86 to 0.88 & 0.88 (18) & 0.72 & 0.85 \\
Llama-3.3-70B & 15 & 0.34 to 0.41 & 0.32 (11) & 0.36 & 0.34 \\
Qwen2.5-72B & 8 & 0.54 to 0.67 & 0.54 (5) & 0.56 & -- \\
\bottomrule
\end{tabular}
\end{table}

\paragraph{What geometry knows.}
Table~\ref{tab:holdout} repeats the test of Sec.~\ref{sec:geometry} on Qwen2.5-7B under three
holdouts, on the matrix behind Table~\ref{tab:predictor}. Behavior names alone match every geometric predictor when single
pairs are held out, which is the shortcut of Sec.~\ref{sec:intro} made concrete. On the best
geometric predictor per model, held-out $R^2$ reaches at most 0.15 on any of the ten models.
Nor does geometry reproduce the average pattern as random pushes do: a behavior's mean cosine to
the other directions has rank correlation $-0.03$ with how easily it is induced on Qwen2.5-7B.

\begin{table}[h]
\centering
\caption{Held-out Spearman / $R^2$ on Qwen2.5-7B under three holdouts. With both behaviors of
a test pair unseen, names carry no information.}
\label{tab:holdout}
\small
\begin{tabular}{lccc}
\toprule
predictor & one pair held out & one behavior held out & both behaviors held out \\
\midrule
behavior names only, no geometry & 0.52 / 0.15 & 0.43 / 0.12 & none / $-$0.03 \\
cosine overlap, ridge & 0.26 / 0.10 & 0.24 / 0.09 & 0.21 / 0.07 \\
all 13 features, ridge & 0.37 / 0.13 & 0.33 / 0.07 & 0.22 / $-$0.06 \\
all 13 features, boosted trees & 0.49 / 0.14 & 0.40 / 0.10 & 0.09 / $-$0.30 \\
\bottomrule
\end{tabular}
\end{table}

\paragraph{The random push at scale.}
Layer-7 pushes enter at 9 to 15 percent of depth on Qwen2.5-72B, Llama-3.3-70B and Gemma-2-27B
against 25 to 34 percent for real steers. Matched pushes (Appendix~\ref{sec:app-new}) remove that
gap and the selection on size. They raise Gemma-2-27B from 0.31 to 0.78 but leave Llama-3.3-70B
at 0.40. On Mistral-7B, Llama-3.1-8B and Gemma-2-9B they agree less than the layer-7 pushes did
(0.36 to 0.51 against 0.72 to 0.80); the layer-7 rule favored the largest size that passed
the bar, and matched pushes at 1.5 times recover 0.55 to 0.81. On all ten models, pushes and real steers alike lower what the
unsteered answer already shows, which is partly arithmetic, because a rating of zero cannot fall.

\paragraph{Steer size.}
On Qwen2.5-7B the median steer adds a vector of norm 48, about 2 times the norm of the mean activation at its layer (1.1 to 3.1), and random
pushes about 3 times (2.0 to 3.9). These use class means, whose norm is smaller
than the norm of single tokens, so they overstate the per-token ratio. Against the median
per-token norm at the steer's layer, measured on unsteered answers, the median steer adds about
1.1 times on Qwen2.5-7B, and the model medians run from 0.5 (Gemma-2-27B) to 1.4 (Qwen2.5-72B).

\paragraph{Single steers and single pairs.}
Table~\ref{tab:headtohead} looks past the average. One steer is the median rank correlation
of a single real steer's side effects with the profile of usable pushes. Splitting the steers
in half by how far they move other behaviors gives the gentle and strong columns, and
splitting by on-target effect gives low and high. The two splits disagree, and the first is
chosen by outcome, so neither settles whether the agreement needs strong steering. The
last two columns score cosine overlap and the random profile, from all 24 pushes, on the same
held-out entries.

\begin{table}[h]
\centering
\caption{Beyond the average. Rank correlations with the random profile for one steer and for
halves of the steers, then held-out $R^2$ on single entries for cosine overlap and the random
profile.}
\label{tab:headtohead}
\small
\begin{tabular}{lccccccc}
\toprule
model & one steer & gentle & strong & low & high & cosine $R^2$ & random $R^2$ \\
\midrule
Qwen2.5-1.5B & 0.62 & $-$0.06 & 0.95 & 0.92 & 0.84 & $-$0.03 & 0.17 \\
Qwen2.5-3B & 0.37 & 0.40 & 0.75 & 0.65 & 0.75 & 0.01 & 0.15 \\
Mistral-7B & 0.26 & 0.50 & 0.68 & 0.72 & 0.56 & 0.12 & 0.09 \\
Qwen2.5-7B & 0.57 & 0.82 & 0.88 & 0.87 & 0.86 & 0.09 & 0.20 \\
Llama-3.1-8B & 0.45 & 0.72 & 0.68 & 0.78 & 0.65 & 0.04 & 0.16 \\
Gemma-2-9B & 0.33 & 0.79 & 0.67 & 0.78 & 0.74 & 0.08 & 0.07 \\
Gemma-2-27B & 0.12 & $-$0.68 & 0.48 & 0.31 & 0.28 & 0.05 & $-$0.00 \\
Qwen2.5-32B & 0.36 & 0.68 & 0.87 & 0.57 & 0.85 & 0.13 & 0.10 \\
Llama-3.3-70B & 0.26 & $-$0.05 & 0.62 & 0.37 & 0.52 & 0.08 & 0.01 \\
Qwen2.5-72B & 0.19 & 0.32 & 0.74 & 0.46 & 0.71 & 0.11 & 0.06 \\
\bottomrule
\end{tabular}
\end{table}

\section{Matched pushes, base models, and further controls}
\label{sec:app-new}

These runs replace the layer-7 random push with matched pushes and add the controls below.
Every run saves its text, and one fixed judge rates every reply afterwards.

\paragraph{Matched pushes.}
For each steerable behavior we draw a random direction with the norm of that behavior's
calibrated steer and add it at the steer's layer, at 0.5, 1 and 1.5 times that size, with two
directions per steer at 1 times. No size is selected after the fact. A push is usable if at
least 80 percent of its 32 generations pass the text checks of Appendix~\ref{sec:app-checks},
in a stricter form that also rejects glued and doubled tokens. Table~\ref{tab:matched} gives
the agreement of usable pushes with published susceptibility. From half to full size it rises
on nine of ten models; beyond that it rises mainly where it was weak. Against susceptibility
rebuilt from clean text only, as in Table~\ref{tab:clean}, the values move by at most 0.15.

\begin{table}[h]
\centering
\caption{Matched pushes. Size is the median real steer's added norm over the median per-token
residual norm at its layer. Usable counts usable pushes out of all pushes at 0.5, 1 and 1.5
times a real steer's size; $\rho$ is their rank correlation with published susceptibility, and
the interval is a 95 percent bootstrap over behaviors at 1 times.}
\label{tab:matched}
\small
\setlength{\tabcolsep}{3pt}
\begin{tabular}{lcccccccc}
\toprule
 & & \multicolumn{3}{c}{usable} & \multicolumn{3}{c}{$\rho$} & \\
\cmidrule(lr){3-5}\cmidrule(lr){6-8}
model & size & $0.5\times$ & $1\times$ & $1.5\times$ & $0.5\times$ & $1\times$ & $1.5\times$ & interval \\
\midrule
Qwen2.5-1.5B & 1.05 & 14/14 & 21/28 & 6/14 & 0.67 & 0.93 & 0.89 & (0.73, 1.00) \\
Qwen2.5-3B & 0.83 & 13/13 & 26/26 & 10/13 & 0.30 & 0.73 & 0.64 & (0.26, 0.94) \\
Mistral-7B & 1.17 & 21/21 & 42/42 & 13/21 & 0.30 & 0.42 & 0.55 & ($-$0.13, 0.85) \\
Qwen2.5-7B & 1.06 & 20/20 & 39/40 & 17/20 & 0.51 & 0.87 & 0.92 & (0.64, 0.96) \\
Llama-3.1-8B & 0.84 & 22/22 & 42/44 & 20/22 & 0.33 & 0.36 & 0.81 & ($-$0.19, 0.81) \\
Gemma-2-9B & 1.20 & 20/20 & 38/40 & 17/20 & 0.39 & 0.51 & 0.72 & (0.07, 0.79) \\
Gemma-2-27B & 0.52 & 19/20 & 21/40 & 5/20 & 0.63 & 0.78 & 0.52 & (0.44, 0.95) \\
Qwen2.5-32B & 1.29 & 22/22 & 39/44 & 18/22 & 0.71 & 0.79 & 0.79 & (0.53, 0.92) \\
Llama-3.3-70B & 1.05 & 24/24 & 48/48 & 20/24 & 0.56 & 0.40 & 0.41 & ($-$0.03, 0.74) \\
Qwen2.5-72B & 1.41 & 22/22 & 44/44 & 21/22 & 0.26 & 0.49 & 0.61 & (0.03, 0.81) \\
\bottomrule
\end{tabular}

\end{table}

\paragraph{Base and instruction-tuned models.}
We compare base and instruction-tuned Qwen2.5-7B, Llama-3.1-8B, Mistral-7B-v0.3 and
Gemma-2-9B. Both versions of a model get the same plain prompt (``User: \ldots\ Assistant:''),
since a base model has no chat template, and one fixed judge, Qwen2.5-72B-Instruct, so the two
differ only in their weights. Directions, calibration, the matrix and the pushes are rebuilt for
each version (Table~\ref{tab:base}). Base models have the same one-way structure, are at least
as low-rank, and are as poorly predicted by geometry. Refusal or sycophancy is among the five
behaviors steers induce most in every instruction-tuned version but in only one base model,
Llama-3.1-8B, and the two versions' orderings agree at 0.71 on Llama-3.1-8B and at $-0.08$ to
0.41 on the other three. Random pushes agree with the base models' ordering at 0.17 to 0.80,
more than with the instruction-tuned versions' in this setting ($-0.05$ to 0.34). That is
lower than in the main text, where these four instruction-tuned models give 0.36 to 0.87 with
their own chat template and judge, and in this setting their orderings agree with their
published ones at only 0.25 to 0.79. The setting changed both the prompt format and the judge,
and we have not separated which of the two matters.

\begin{table}[h]
\centering
\caption{Base and instruction-tuned models on one plain prompt format with one fixed judge.
Steer.\ is the number of steerable behaviors, one-way the antisymmetric norm fraction as in
Table~\ref{tab:scaling}, top-3 the energy share of the top three patterns, geom.\ $R^2$ the
held-out $R^2$ of the 13-feature ridge predictor, and push $\rho$ the agreement of random pushes
with that version's susceptibility. The last column lists the five behaviors steers induce
most, in order.}
\label{tab:base}
\small
\setlength{\tabcolsep}{3pt}
\begin{tabular}{llccccc>{\raggedright\arraybackslash}p{4.6cm}}
\toprule
model & version & steer. & one-way & top-3 & geom.\ $R^2$ & push $\rho$ & five most induced \\
\midrule
Qwen2.5-7B & base & 14 & 0.55 & 0.81 & 0.17 & 0.48 & formality, sentiment, politeness, bureaucratese, poeticism \\
 & instruct & 19 & 0.61 & 0.73 & 0.10 & 0.27 & refusal, poeticism, hedging, bureaucratese, archaism \\
\addlinespace[2pt]
Llama-3.1-8B & base & 18 & 0.69 & 0.70 & 0.02 & 0.80 & verbosity, poeticism, refusal, pessimism, hedging \\
 & instruct & 21 & 0.66 & 0.68 & 0.11 & 0.34 & verbosity, archaism, poeticism, sycophancy, pessimism \\
\addlinespace[2pt]
Mistral-7B & base & 15 & 0.63 & 0.81 & 0.03 & 0.17 & archaism, formality, optimism, sentiment, bureaucratese \\
 & instruct & 18 & 0.62 & 0.72 & 0.17 & $-$0.05 & politeness, poeticism, empathy, sycophancy, archaism \\
\addlinespace[2pt]
Gemma-2-9B & base & 18 & 0.62 & 0.79 & 0.15 & 0.76 & optimism, sentiment, verbosity, confidence, formality \\
 & instruct & 20 & 0.62 & 0.73 & 0.20 & 0.26 & verbosity, poeticism, refusal, hedging, bureaucratese \\
\bottomrule
\end{tabular}

\end{table}

\paragraph{Gentler and stronger steers.}
On five models we rebuild the matrix with every steer at 0.5 to 1.5 times its calibrated
coefficient, with one matched push per steer at the same multiple (Table~\ref{tab:strength}).
The one-way share stays between 0.57 and 0.72 at every strength, and at half strength the top
three patterns carry more of the energy, not less. The susceptibility ordering agrees with the
published one at 0.81 or more from 0.75 times upward, and at 0.59 to 0.84 at half strength.
Measured against the matrix at the same strength, push agreement at half strength is at least
as high as at full strength on four of the five models, and lower only on Qwen2.5-1.5B (0.48
against 0.89), so it does not need strong steering on those four.

\begin{table}[h]
\centering
\caption{The matrix at 0.5 to 1.5 times the calibrated strength. One-way and top-3 as in
Table~\ref{tab:base}; order is the rank correlation of susceptibility with the published one;
usable text is the share of real-steer generations that pass the text checks; push $\rho$ is the
agreement of usable pushes at that strength with the matrix at that strength.}
\label{tab:strength}
\small
\setlength{\tabcolsep}{4pt}
\begin{tabular}{llccccc}
\toprule
model & quantity & $0.5\times$ & $0.75\times$ & $1\times$ & $1.25\times$ & $1.5\times$ \\
\midrule
Qwen2.5-1.5B & one-way & 0.69 & 0.68 & 0.68 & 0.66 & 0.69 \\
 & top-3 & 0.99 & 0.92 & 0.93 & 0.91 & 0.91 \\
 & order & 0.73 & 0.96 & 1.00 & 0.93 & 0.86 \\
 & usable text & 1.00 & 0.96 & 0.87 & 0.78 & 0.70 \\
 & push $\rho$ & 0.48 & 0.95 & 0.89 & 0.93 & 0.98 \\
\addlinespace[2pt]
Qwen2.5-7B & one-way & 0.70 & 0.68 & 0.67 & 0.64 & 0.61 \\
 & top-3 & 0.84 & 0.78 & 0.71 & 0.72 & 0.75 \\
 & order & 0.62 & 0.81 & 0.99 & 0.96 & 0.96 \\
 & usable text & 1.00 & 1.00 & 0.95 & 0.78 & 0.56 \\
 & push $\rho$ & 0.90 & 0.82 & 0.82 & 0.85 & 0.88 \\
\addlinespace[2pt]
Llama-3.1-8B & one-way & 0.69 & 0.72 & 0.66 & 0.66 & 0.68 \\
 & top-3 & 0.93 & 0.74 & 0.67 & 0.68 & 0.70 \\
 & order & 0.62 & 0.91 & 1.00 & 0.95 & 0.92 \\
 & usable text & 0.94 & 0.94 & 0.91 & 0.85 & 0.77 \\
 & push $\rho$ & 0.47 & 0.53 & 0.38 & 0.53 & 0.75 \\
\addlinespace[2pt]
Gemma-2-9B & one-way & 0.70 & 0.68 & 0.68 & 0.63 & 0.65 \\
 & top-3 & 0.78 & 0.75 & 0.68 & 0.75 & 0.79 \\
 & order & 0.84 & 0.94 & 0.99 & 0.96 & 0.92 \\
 & usable text & 0.98 & 0.97 & 0.92 & 0.75 & 0.65 \\
 & push $\rho$ & 0.58 & 0.55 & 0.54 & 0.58 & 0.50 \\
\addlinespace[2pt]
Llama-3.3-70B & one-way & 0.66 & 0.67 & 0.63 & 0.57 & 0.58 \\
 & top-3 & 0.73 & 0.85 & 0.70 & 0.65 & 0.64 \\
 & order & 0.59 & 0.91 & 1.00 & 0.97 & 0.90 \\
 & usable text & 1.00 & 1.00 & 0.98 & 0.87 & 0.77 \\
 & push $\rho$ & 0.79 & 0.34 & 0.35 & 0.44 & 0.53 \\
\bottomrule
\end{tabular}

\end{table}

\paragraph{Matched on-target strength.}
For each behavior we search its coefficient by bisection, starting from the published sweep,
until its own judged effect is 2 points with a tolerance of 0.5, on text of which at least 80
percent is usable, and rebuild the matrix at those coefficients (Table~\ref{tab:matchedself}).
On Qwen2.5-7B 17 of 20 behaviors reach the tolerance, and source strength and susceptibility
keep rank correlations of 0.87 and 0.95 with the published rankings. Across models source
strength keeps 0.66 to 0.96 and susceptibility 0.71 to 0.99, except susceptibility on
Qwen2.5-3B (0.48, from 9 behaviors within tolerance). Unequal on-target strength does not
produce the ranking.

\begin{table}[h]
\centering
\caption{Every behavior at the same on-target effect. Within counts behaviors whose effect lies
within 0.5 of 2 judge points; coef.\ is the median ratio of the chosen coefficient to the
published one; the last two columns are rank correlations with the published rankings.}
\label{tab:matchedself}
\small
\begin{tabular}{lcccc}
\toprule
model & within & coef. & source strength & susceptibility \\
\midrule
Qwen2.5-1.5B & 10/14 & 1.00 & 0.96 & 0.99 \\
Qwen2.5-3B & 9/13 & 1.00 & 0.75 & 0.48 \\
Mistral-7B & 19/21 & 0.88 & 0.91 & 0.93 \\
Qwen2.5-7B & 17/20 & 0.93 & 0.87 & 0.95 \\
Llama-3.1-8B & 21/22 & 0.88 & 0.82 & 0.93 \\
Gemma-2-9B & 17/20 & 0.98 & 0.72 & 0.96 \\
Gemma-2-27B & 19/20 & 1.00 & 0.66 & 0.71 \\
Qwen2.5-32B & 19/22 & 0.88 & 0.81 & 0.95 \\
Llama-3.3-70B & 21/24 & 1.00 & 0.94 & 0.99 \\
Qwen2.5-72B & 20/22 & 0.92 & 0.70 & 0.86 \\
\bottomrule
\end{tabular}

\end{table}

\paragraph{Pushes through the layers.}
On each model we add 24 random pushes at the median real steer's layer and size and follow the
change each makes to the residual stream at the last prompt token, with the model reading its
own unsteered answer (Table~\ref{tab:trace}). The 24 changes start unrelated, with effective
dimension about 24 and mean cosine 0. By the last layer before the final norm they share one
direction on nine of ten models: effective dimension 1.2 to 4.7 and mean cosine 0.41 to 0.91.
Gemma-2-9B converges less. On those nine the shared change points the way real steers' changes
do (cosine 0.38 to 0.58), but its alignment with each behavior's direction does not predict
which behaviors pushes induce (rank correlation $-0.14$ to 0.49). No model but Gemma-2-9B
absorbs the push: on the other nine the change at the last layer is 1.2 to 7.5 times what was
added, with no trend toward shrinking at scale.

\begin{table}[h]
\centering
\caption{Random pushes followed through the layers, at the last prompt token. Layers runs from
the injection layer to the last layer before the final norm. Dimension is the participation
ratio of the 24 changes, first to last; cosine is their mean pairwise cosine at the last layer;
real is the cosine of their mean change with real steers' changes; align.\ is the rank
correlation of the mean change's alignment with each behavior's direction with what pushes
induce; growth is the size of the change at the last layer over the size added.}
\label{tab:trace}
\small
\setlength{\tabcolsep}{4pt}
\begin{tabular}{lcccccc}
\toprule
model & layers & dimension & cosine & real & align. & growth \\
\midrule
Qwen2.5-1.5B & 7 to 27 & 23.7 to 3.2 & 0.52 & 0.48 & 0.24 & 2.06 \\
Qwen2.5-3B & 9 to 35 & 23.8 to 3.9 & 0.45 & 0.38 & 0.21 & 1.16 \\
Mistral-7B & 8 to 31 & 23.9 to 4.7 & 0.41 & 0.38 & 0.49 & 2.25 \\
Qwen2.5-7B & 7 to 27 & 23.9 to 2.2 & 0.64 & 0.58 & 0.33 & 2.85 \\
Llama-3.1-8B & 8 to 31 & 23.9 to 3.4 & 0.48 & 0.47 & 0.29 & 2.58 \\
Gemma-2-9B & 10 to 41 & 23.9 to 7.1 & 0.30 & 0.20 & $-$0.13 & 0.61 \\
Gemma-2-27B & 15 to 45 & 23.9 to 1.2 & 0.91 & 0.48 & $-$0.14 & 7.52 \\
Qwen2.5-32B & 16 to 63 & 23.9 to 1.9 & 0.68 & 0.51 & 0.34 & 1.49 \\
Llama-3.3-70B & 20 to 79 & 23.9 to 2.1 & 0.64 & 0.38 & $-$0.01 & 2.15 \\
Qwen2.5-72B & 20 to 79 & 23.9 to 1.7 & 0.75 & 0.46 & 0.32 & 4.71 \\
\bottomrule
\end{tabular}

\end{table}

\paragraph{The concept cone.}
Built as in Appendix~\ref{sec:app-directions}, so that it contains the difference of means, the
cone adds nothing to cosine overlap on Qwen2.5-7B: held-out $R^2$ is 0.072 for the two together
against 0.090 for cosine alone, and 0.009 for the cone alone (Table~\ref{tab:predictor}).

\paragraph{What the judge scores as refusal, and fluent text.}
One fixed judge, Qwen2.5-72B-Instruct, reads each distinct question and reply, 104,570 in all,
and rates whether the reply is fluent (0 to 10), whether it is on topic, how far it answers
(0 to 10), and whether it explicitly declines. Of the 509 side-effect generations scored 5 or
more on refusal, 97 percent answer little or nothing (3 or less) and 32 percent explicitly
decline; 23 percent of the non-answers are fluent (7 or more). What the judge scores as refusal
is therefore almost always a failure to answer, and about a third of the time a refusal in words.
Keeping only fluent generations leaves 5 to 21 behaviors per model with enough text. On those,
the susceptibility ordering from fluent text agrees with the ordering from all text at 0.87 to
1.00 on all ten models. The fluency rating also reacts to tone, rating rude or ranting replies
lower, so this filter removes some style along with broken text.

\end{document}